\PassOptionsToPackage{unicode}{hyperref}
\PassOptionsToPackage{hyphens}{url}
\PassOptionsToPackage{dvipsnames,svgnames,x11names}{xcolor}
\documentclass[
  10pt]{article}
\usepackage{amsmath,amssymb}
\usepackage{lmodern}
\usepackage{iftex}
\ifPDFTeX
  \usepackage[T1]{fontenc}
  \usepackage[utf8]{inputenc}
  \usepackage{textcomp} 
\else 
  \usepackage{unicode-math}
  \defaultfontfeatures{Scale=MatchLowercase}
  \defaultfontfeatures[\rmfamily]{Ligatures=TeX,Scale=1}
\fi
\IfFileExists{upquote.sty}{\usepackage{upquote}}{}
\IfFileExists{microtype.sty}{
  \usepackage[]{microtype}
  \UseMicrotypeSet[protrusion]{basicmath} 
}{}
\makeatletter
\@ifundefined{KOMAClassName}{
  \IfFileExists{parskip.sty}{%
    \usepackage{parskip}
  }{
    \setlength{\parindent}{0pt}
    \setlength{\parskip}{6pt plus 2pt minus 1pt}}
}{
  \KOMAoptions{parskip=half}}
\makeatother
\usepackage{xcolor}
\usepackage{longtable,booktabs,array}
\usepackage{calc} 
\usepackage{etoolbox}
\makeatletter
\patchcmd\longtable{\par}{\if@noskipsec\mbox{}\fi\par}{}{}
\makeatother
\usepackage{graphicx}
\makeatletter
\def\maxwidth{\ifdim\Gin@nat@width>\linewidth\linewidth\else\Gin@nat@width\fi}
\def\maxheight{\ifdim\Gin@nat@height>\textheight\textheight\else\Gin@nat@height\fi}
\makeatother
\setkeys{Gin}{width=\maxwidth,height=\maxheight,keepaspectratio}
\makeatletter
\def\fps@figure{htbp}
\makeatother
\providecommand{\tightlist}{%
  \setlength{\itemsep}{0pt}\setlength{\parskip}{0pt}}
\newlength{\cslhangindent}
\newlength{\csllabelwidth}
\newlength{\cslentryspacingunit} 
\newenvironment{CSLReferences}[2] 
 {
  \setlength{\parindent}{0pt}
  \ifodd #1
  \let\oldpar\par
  \def\par{\hangindent=\cslhangindent\oldpar}
  \fi
  \setlength{\parskip}{#2\cslentryspacingunit}
 }%
 {}
\usepackage{calc}

\usepackage{fullpage}
\usepackage{setspace}
\usepackage{parskip}
\usepackage{titlesec}
\usepackage[section]{placeins}
\usepackage{breakcites}
\usepackage{lineno}
\usepackage{hyphenat}
\usepackage[all]{nowidow}

\usepackage{etoolbox}

\renewenvironment{abstract}
  {{\bfseries\noindent{\abstractname}\par\nobreak}\footnotesize}
  {\bigskip}
\titlespacing{\section}{0pt}{*3}{*1}
\titlespacing{\subsection}{0pt}{*2}{*0.5}
\titlespacing{\subsubsection}{0pt}{*1.5}{0pt}

\usepackage{authblk}

\usepackage{graphicx}
\usepackage{enumitem}
\usepackage{tikz}
\usetikzlibrary{matrix,arrows,decorations.pathreplacing}
\tikzstyle{every node}=[font=\fontsize{8}{10}\selectfont]

\usepackage[space]{grffile}
\usepackage{latexsym}
\usepackage{textcomp}
\usepackage{longtable}
\usepackage{tabulary}
\usepackage{booktabs,array,multirow}
\usepackage{amsfonts,amsmath,amssymb}
\usepackage{subcaption}
\providecommand\citet{\cite}
\providecommand\citep{\cite}

\newif\iflatexml\latexmlfalse
\providecommand{\tightlist}{\setlength{\itemsep}{0pt}\setlength{\parskip}{0pt}}

\AtBeginDocument{\DeclareGraphicsExtensions{.pdf,.PDF,.eps,.EPS,.png,.PNG,.tif,.TIF,.jpg,.JPG,.jpeg,.JPEG}}

\usepackage[utf8]{inputenc}
\usepackage[english]{babel}
\usepackage{float}

\usepackage[margin=1.5in]{geometry}
\ifdefined\N                                                                
\renewcommand{\N}{\mathds{N}}                                                
\else
  \newcommand{\N}{\mathds{N}}
\fi
\ifdefined\C
  \renewcommand{\C}{\mathds{C}}                                             
\else
  \newcommand{\C}{\mathds{C}}
\fi

\renewcommand{\xi}[1][i]{\mathbf{x}^{(#1)}}                                          

\newcommand{\hdspace}{\mathcal{H}}

\newcommand{\embedspace}{\mathcal{Y}}
\newcommand{\pspace}{\Theta}
\newcommand{\mani}{\mathcal{M}}

\newcommand{\vizdim}{\mathbf{D}}        
\newcommand{\obsdim}{\ensuremath{D}}    

\newcommand{\co}{c}
\newcommand{\an}{a}
\newcommand{\Min}{\mathcal{M}_{\co}}
\newcommand{\Man}{\mathcal{M}_{\an}}
\newcommand{\Pin}{P_{\co}}
\newcommand{\Pan}{P_{\an}}
\newcommand{\phin}{\phi_{\co}}
\newcommand{\phan}{\phi_{\an}}
\newcommand{\Thin}{\Theta_{\co}}
\newcommand{\Than}{\Theta_{\an}}

\usepackage{float}
\usepackage{hyperref}
\IfFileExists{bookmark.sty}{\usepackage{bookmark}}{\usepackage{hyperref}}
\hypersetup{
  colorlinks=true,
  linkcolor={blue},
  filecolor={Maroon},
  citecolor={Blue},
  urlcolor={Blue},
  pdfcreator={LaTeX via pandoc}}

\author{}
\date{\vspace{-2.5em}}

\begin{document}

\title{A geometric framework for outlier detection in high-dimensional data}

\def\correspondingauthor{\footnote{Corresponding author, e-mail: \href{mailto:moritz.herrmann@stat.uni-muenchen.de}{moritz.herrmann@stat.uni-muenchen.de}, Department of Statistics, Ludwig Maximilians University Munich, Ludwigstr. 33, D-80539, Munich, Germany.}}
\author{Moritz Herrmann\correspondingauthor, Florian Pfisterer, and Fabian Scheipl \\
Department of Statistics, Ludwig Maximilians University, Munich, Germany}

\vspace{-1em}
  \date{}
\begingroup
\let\center\flushleft
\let\endcenter\endflushleft
\maketitle
\endgroup
\selectlanguage{english}

\vspace{1cm}

\begin{abstract}
Outlier or anomaly detection is an important task in data analysis. We discuss the problem from a geometrical perspective and provide a framework which exploits the metric structure of a data set. Our approach rests on the \textit{manifold assumption}, i.e., that the observed, nominally high-dimensional data lie on a much lower dimensional manifold and that this intrinsic structure can be inferred with manifold learning methods. We show that exploiting this structure significantly improves the detection of outlying observations in high dimensional data. We also suggest a novel, mathematically precise and widely applicable distinction between \textit{distributional} and \textit{structural} outliers based on the geometry and topology of of the data manifold that clarifies conceptual ambiguities prevalent throughout the literature.
Our experiments focus on functional data as one class of structured high-dimensional data, but the framework we propose is completely general and we include image and graph data applications. Our results show that the outlier structure of high-dimensional and non-tabular data can be detected and visualized using manifold learning methods and quantified using standard outlier scoring methods applied to the manifold embedding vectors.
\end{abstract}

\hypertarget{sec:intro}{%
\section{Introduction}\label{sec:intro}}

Detecting atypical observations that deviate substantially from the bulk
of the data is an important task in data analysis with applications
across domains like, e.g., intrusion detection
(\protect\hyperlink{ref-zhang2006anomaly}{Zhang \& Zulkernine, 2006}),
medical imaging (\protect\hyperlink{ref-fritsch2012detecting}{Fritsch et
al., 2012}), or network analysis
(\protect\hyperlink{ref-azcorra2018unsupervised}{Azcorra et al., 2018}).
The most common terms for this task are \emph{outlier} or \emph{anomaly
detection}, but many different terms are used
(\protect\hyperlink{ref-zimek2018there}{Zimek \& Filzmoser, 2018}).
Although there is a vast amount of literature on the topic, there is
neither a commonly accepted, precise definition of what exactly
constitutes outliers or anomalies, nor agreement on whether these two
terms are synonymous. As Unwin
(\protect\hyperlink{ref-unwin2019multivariate}{2019, p. 635}) puts it:

\begin{quote}
``Outliers are a complicated business. It is difficult to define what
they are, it is difficult to identify them, and it is difficult to
assess how they affect analyses.''
\end{quote}

\noindent Overviews on the topic are given by Zimek et al.
(\protect\hyperlink{ref-zimek2012survey}{2012}) or Goldstein \& Uchida
(\protect\hyperlink{ref-goldstein2016comparative}{2016}) from a computer
science perspective, and by Rousseeuw \& Leroy
(\protect\hyperlink{ref-rousseeuw2005robust}{2005}) or Unwin
(\protect\hyperlink{ref-unwin2019multivariate}{2019}) from a statistical
perspective. Kandanaarachchi \& Hyndman
(\protect\hyperlink{ref-kandanaarachchi2020dimension}{2020}) provide a
short summary including both perspectives, while Campos et al.
(\protect\hyperlink{ref-campos2016evaluation}{2016}) as well as Marques
et al. (\protect\hyperlink{ref-marques2020internal}{2020}) focus on the
evaluation of unsupervised outlier detection. Zimek \& Filzmoser
(\protect\hyperlink{ref-zimek2018there}{2018}) provide a comprehensive
survey bringing together both perspectives with in-depth epistemological
discussion. \color{black} In particular, Zimek \& Filzmoser
(\protect\hyperlink{ref-zimek2018there}{2018}) discuss that there are
two different notions of outliers and different terms used to describe
these notions -- including, for example, \emph{apparent},
\emph{discrepant}, \emph{real}, \emph{contaminating}, or \emph{true}
outlier -- in the literature. From this discussion, it can be inferred
that (1) Zimek \& Filzmoser
(\protect\hyperlink{ref-zimek2018there}{2018, p. 7}) distinguish
``true'' and ``apparent'' outliers and consider ``those objects as
`(true) outliers' that have been `generated by a different mechanism'
than the remainder or major part of the data or than the whatsoever
defined reference set'', (2) there is neither a clear understanding of
how these two notions are different and actually manifest in practice
nor (3) a ``language'' to precisely describe the problem
theoretically.\\
In the more statistically flavored literature, the problem of
unsupervised outlier detection is usually tackled by defining outliers
based on a single probability distribution \(P\). If \(P\) allows for a
density, outliers are simply observations in low-density regions. From
this perspective, we have \textit{distributional outliers} whose
outlyingness is defined relative to a single probability distribution.
The notion of \emph{distributional outliers} is easy to define precisely
in probabilistic terms, for example, based on minimum level sets
(\protect\hyperlink{ref-scott2006learning}{Scott \& Nowak, 2006}) or
M-estimation (\protect\hyperlink{ref-clemenccon2013scoring}{Clémençon \&
Jakubowicz, 2013}), and has yielded a multitude of results and
algorithms. In practical terms, this requires access to (an estimate of)
the underlying density and finding a suitable (local) density level
below which observations are to be classified as outliers. Note that
both are infeasible for general, non-tabular data types like shapes,
functions, or images whose domains frequently do not admit probability
densities. However, Zimek \& Filzmoser
(\protect\hyperlink{ref-zimek2018there}{2018}) emphasize that
``observations which are in the extremes of the model distribution
{[}i.e., distributional outlier{]} should be distinguished from `real'
outliers (contaminants)'' (\protect\hyperlink{ref-zimek2018there}{Zimek
\& Filzmoser, 2018, p. 13}). This second notion of outliers (``true'' or
``real'' outliers) is not reflected by the statistical concept because
such outliers are assumed to be observations generated by a different
data-generating process. This is reflected in statements like
``different mechanism'' or ``any observation that is not a realization
from the target distribution''
(\protect\hyperlink{ref-beckmanOutlier1983}{Beckman \& Cook, 1983, p.
121}). That means, for the second notion (``real outliers'') it is
implicitly assumed that outliers are not independent and identically
distributed (IID) observations. So next to
\textit{distributional outliers} there are also
\textit{structural outliers} whose outlyingness is caused by the
structural differences between the underlying data generating processes.
The two outlier types are complementary and both are necessary to fully
address the challenges of outlier detection. In contrast to
\emph{distributional outliers}, \emph{structural outliers} are much more
difficult to formalize, but also more general.\\
With this work, we intend to ``broaden'' the view on the problem of
unsupervised outlier detection to account for the two notions of
outliers present in the literature. We show that a \emph{geometric}
approach to the problem, which does not require the availability of
probability densities defined over the data space but only some metric
structure (i.e., suitable dissimilarity or distance measures), allows
for a more precise conceptualization and terminology. To do so, we focus
on building up intuition and demonstrating the application of these
concepts to diverse and comprehensive practical examples and
visualizations. However, to be able to ``speak'' of this new perspective
without referring to vague and subjective perceptions as done previously
(see \protect\hyperlink{ref-zimek2018there}{Zimek \& Filzmoser, 2018}),
we also consider it necessary to introduce a certain degree of
mathematical terminology. The provided degree of formality is exhausted
in the definition of \emph{distributional} and \emph{structural}
outliers in precise mathematical terms and thus serves the need for
precise terminology but does not overload the work with more formalism
than we think necessary to contribute to the overall scope of the study.
Readers interested in more rigorous mathematical approaches to infer
structures in data may, for example, consult Mordohai \& Medioni
(\protect\hyperlink{ref-mordohaiDimensionalityEstimationManifold2010}{2010})
for dimensionality estimation and manifold learning based on tensor
voting, Niyogi et al.
(\protect\hyperlink{ref-niyogiTopologicalViewUnsupervised2011}{2011})
for a topological perspective on unsupervised learning, Guan \& Loew
(\protect\hyperlink{ref-guanNovelIntrinsicMeasure2021}{2021}) of a
distance-based measure of class separability, and Kandanaarachchi \&
Hyndman (\protect\hyperlink{ref-kandanaarachchi2020dimension}{2020}) for
outlier detection in tabular data based on dimensionality reduction.\\
\color{black}

The rest of the paper is structured as follows. Section
\ref{sec:prelims} describes the scope and contribution of the study and
outlines its background and related work. The proposed theoretical
framework is defined in section \ref{sec:framework} and its practical
relevance is demonstrated in section \ref{sec:exps} using qualitative
and quantitative experiments for a variety of data sets of different
data types. Section \ref{sec:discussion} discusses our findings and the
resulting conceptual implications, before we conclude in section
\ref{sec:conclusion}.

\hypertarget{sec:prelims}{%
\section{Preliminaries}\label{sec:prelims}}

\hypertarget{sec:prelims:scope}{%
\subsection{Scope and contribution of the
study}\label{sec:prelims:scope}}

\color{black}

With this focus article, we intend to draw connections between different
conceptual aspects provided in the overview article by Zimek \&
Filzmoser (\protect\hyperlink{ref-zimek2018there}{2018}) and as proposed
by us in a paper focusing on functional data analysis (FDA). Therefore,
we recapitulate the underlying conceptualization presented in the
earlier paper in a more general form and different terminology in
Section 3. This framework builds on principles from \emph{manifold
learning} (\protect\hyperlink{ref-lee2007nonlinear}{Lee \& Verleysen,
2007}; \protect\hyperlink{ref-ma2011manifold}{Ma \& Fu, 2011}), i.e.,
dimension reduction methods that infer the intrinsic lower-dimensional
manifold structure of high-dimensional data and yield low-dimensional
vector representations of the data. This perspective allows us to
formalize structural and distributional outliers jointly in a single
mathematical framework, where structural outliers are data that are
separate from the main data manifold, and distributional outliers are
data that are situated at the periphery of, but still on the main data
manifold. While the first paper exclusively focused on the functional
data setting, the present focus article generalizes the underlying
conceptualization of outlier detection to other data types. This is
straightforward theoretically but has important general conceptual
implications that have never been described in detail and demonstrated
on diverse real data problems before. In particular, we draw connections
between and provide a unifying perspective on different data types
(functions, images, graphs, tabular data) which were previously often
treated separately from a theoretical as well as a practical
perspective, in particular when it comes to outlier detection.\\
The main contribution of this review paper is to discuss and demonstrate
two conceptual aspects of outlier detection in general. First of all, as
already outlined, there seems to be a lack of clarity about what defines
outliers, evidenced also by the plethora of terms used to describe the
issue (\protect\hyperlink{ref-zimek2018there}{Zimek \& Filzmoser,
2018}). Several recent reviews on the topic also point out this
conceptual ambiguity
(\protect\hyperlink{ref-goldstein2016comparative}{Goldstein \& Uchida,
2016}; \protect\hyperlink{ref-unwin2019multivariate}{Unwin, 2019};
\protect\hyperlink{ref-zimek2018there}{Zimek \& Filzmoser, 2018}). In
particular, the comprehensive overview of Zimek and Filzmoser
(\protect\hyperlink{ref-zimek2018there}{2018, p. 4}) devotes a complete
section to the question of ``what an `outlier' possibly means''. Recall
that they define ``true outliers'' as objects ``that have been
`generated by a different mechanism' than the remainder or major part of
the data or than the whatsoever defined reference set'' and distinguish
them from ``objects that appear to be outliers (independent of whether
or not they actually are (true) outliers'')
(\protect\hyperlink{ref-zimek2018there}{Zimek \& Filzmoser, 2018, p.
7}). As we will show, the geometrical framework provides suitable
mathematical terminology to delineate ``true'' and ``apparent'' outliers
much more cleanly and thus reduces the conceptual ambiguity that
surrounds the topic: We transfer concepts established in manifold
learning to the problem of outlier detection, deriving a novel
underlying conceptualization of the problem of outlier detection that
(1) is capable of reflecting two types of outliers, (2) replaces vague
notions of ``real'', ``contaminant'', or ``apparent'' outliers with a
precise definition, (3) incorporates the well-established concept of
\emph{distributional} outliers in a unified fashion. With this, we can
abandon vague notions of outlier subtypes in favor of two precisely
defined concepts.\\
Second, our framework also suggests that outlier detection in
high-dimensional (and/or non-tabular) data is not necessarily more
challenging than in low-dimensional settings once the underlying
manifold structure is recovered and exploited. This is important because
high dimensionality is often reported to be particularly problematic for
outlier detection and many outlier detection methods break down or at
least face particular challenges in such settings
(\protect\hyperlink{ref-aggarwal2017outlier}{Aggarwal, 2017};
\protect\hyperlink{ref-aggarwal2001outlier}{Aggarwal \& Yu, 2001};
\protect\hyperlink{ref-goldstein2016comparative}{Goldstein \& Uchida,
2016}; \protect\hyperlink{ref-kamalov2020outlier}{Kamalov \& Leung,
2020}; \protect\hyperlink{ref-navarro2021high}{Navarro-Esteban \&
Cuesta-Albertos, 2021}; \protect\hyperlink{ref-ro2015outlier}{Ro et al.,
2015}; \protect\hyperlink{ref-thudumu2020comprehensive}{Thudumu et al.,
2020}; \protect\hyperlink{ref-xu2018comparison}{Xu et al., 2018};
\protect\hyperlink{ref-zimek2012survey}{Zimek et al., 2012}, e.g.).\\
To highlight these aspects, we again provide simple and easily
accessible functional data examples to demonstrate the principal
practical implications (in addition to the recapitulation of the
theoretical conceptualization). Functional data analysis
(\protect\hyperlink{ref-ramsay2005functional}{Ramsay \& Silverman,
2005}, e.g.) deals with data that are (discretized) realizations of
stochastic processes over a compact domain. Functional data is well
suited to illustrate the underlying conceptualization both practically
and theoretically because it is usually highly structured (the manifold
assumption is specifically realistic and useful),
theoretically/analytically well accessible, and easily visualized in
bulk. Beyond the FDA setting, we also use examples of other data types
including image, graph, curve, and tabular data.\\
Finally, our framework is fully general and does not rely on a specific
combination of manifold learning and outlier detection methods. To
demonstrate its practical performance, we show that one of the simplest
and most established manifold learning methods -- Multidimensional
Scaling (MDS) (\protect\hyperlink{ref-cox2008multidimensional}{Cox \&
Cox, 2008}) -- combined with a standard outlier detection algorithm --
Local Outlier Factors (LOF)
(\protect\hyperlink{ref-breunig2000lof}{Breunig et al., 2000}) --
already yields a flexible, reliable, and generally applicable recipe for
outlier detection and visualization in complex, high-dimensional data.
\color{black}

\hypertarget{sec:prelims:background}{%
\subsection{Background and related work}\label{sec:prelims:background}}

The fundamental assumption of manifold learning is that the
high-dimensional data observed in a \(\obsdim\)-dimensional space
\(\hdspace\) actually lie on a \(d\)-dimensional manifold
\(\mani \subset \hdspace\) with \(d < \obsdim\). Manifold learning
methods yield an \emph{embedding} function
\(e:\hdspace \to \embedspace\) from the high-dimensional data space to a
low-dimensional embedding space \(\embedspace\) such that the
configuration of embedded data reflects the characteristics of
\(\mani\). The terms manifold learning and nonlinear dimension reduction
are often used interchangeably
(\protect\hyperlink{ref-lee2007nonlinear}{Lee \& Verleysen, 2007};
\protect\hyperlink{ref-ma2011manifold}{Ma \& Fu, 2011}). Typically, the
fundamental step is to compute distances between the high-dimensional
observations. Methods based on this approach are, for example,
Multidimensional Scaling (MDS)
(\protect\hyperlink{ref-cox2008multidimensional}{Cox \& Cox, 2008}),
Isomap (\protect\hyperlink{ref-tenenbaum2000global}{Tenenbaum et al.,
2000}), diffusion maps
(\protect\hyperlink{ref-coifman2006diffusion}{Coifman \& Lafon, 2006}),
local linear embeddings
(\protect\hyperlink{ref-roweis2000nonlinear}{Roweis \& Saul, 2000}),
Laplacian eigenmaps (\protect\hyperlink{ref-belkin2003laplacian}{Belkin
\& Niyogi, 2003}), t-distributed stochastic neighborhood embeddings
(t-SNE) (\protect\hyperlink{ref-maaten2008visualizing}{Maaten \& Hinton,
2008}), and uniform manifold approximation and projection (UMAP)
(\protect\hyperlink{ref-mcinnes2018umap}{McInnes et al., 2018}), to name
only a few. The methods differ in how they infer the manifold structure
from these distances and how they obtain low-dimensional embedding
vectors from these.\\
Despite their promising results in other settings, manifold learning
methods have not found application for outlier detection to a
significant extent so far. Kandanaarachchi \& Hyndman
(\protect\hyperlink{ref-kandanaarachchi2020dimension}{2020}) define an
outlier detection method explicitly based on dimension reduction, while
Pang et al. (\protect\hyperlink{ref-pang2018learning}{2018}) make use of
ranking model-based representation learning. However, they do not
provide a general conceptual framework and focus on tabular data. For
functional data, Xie et al.
(\protect\hyperlink{ref-xie2017geometric}{2017}) introduce a geometric
approach that decomposes functional observations into amplitude, phase,
and shift components in order to identify specific types of outliers.
However, the approach is only applicable to functional data and does not
make use of the intrinsic structure of the functional observations from
a manifold learning perspective. Ali et al.
(\protect\hyperlink{ref-ali2019timecluster}{2019}) analyze time series
data using 2\(\vizdim\)-embeddings obtained from manifold methods for
outlier detection and clustering and Toivola et al.
(\protect\hyperlink{ref-toivola2010novelty}{2010}) compare specific
dimensionality reduction techniques for outlier detection in structural
health monitoring, but both focus on practical considerations and do not
provide a theoretical underpinning. Another line of work focuses on
projection-based outlier detection, for example for high-dimensional
Gaussian data (\protect\hyperlink{ref-navarro2021high}{Navarro-Esteban
\& Cuesta-Albertos, 2021}), financial time series
(\protect\hyperlink{ref-loperfido2020kurtosis}{Loperfido, 2020}), or
functional data (\protect\hyperlink{ref-ren2017projection}{Ren et al.,
2017}).

\hypertarget{sec:framework}{%
\section{Geometrical framework for outlier
detection}\label{sec:framework}}

\begin{figure}
\centering
\includegraphics{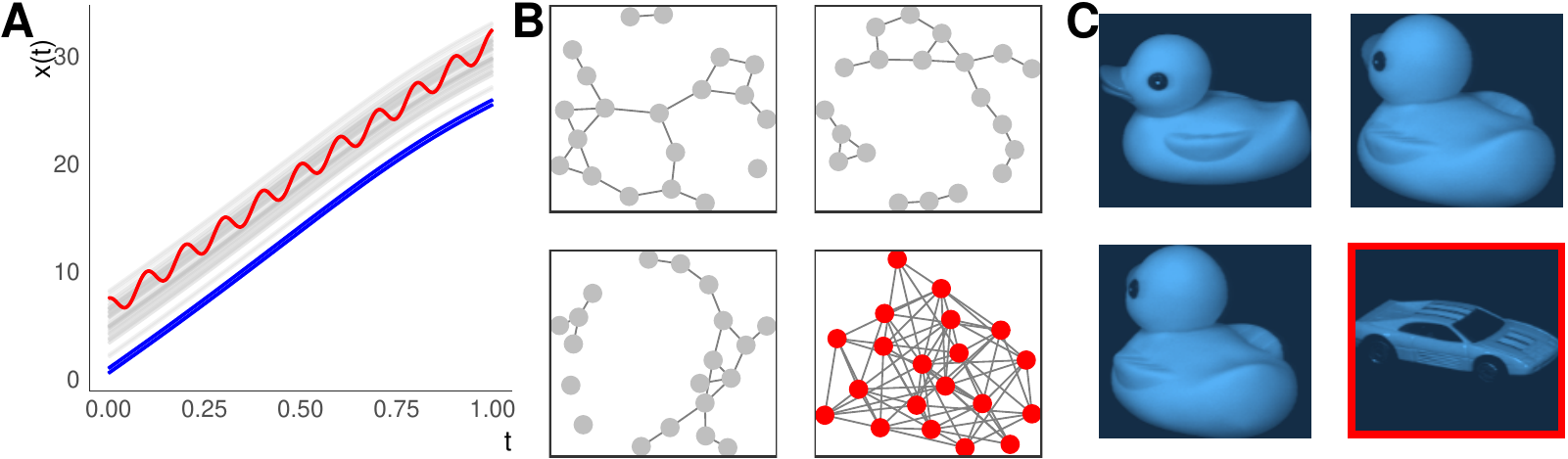}
\caption{\label{fig:outtypes}Example data types. A: Functional inliers
(grey) with a structural outlier (red) and distributional outliers
(blue). B: Graph data with a structural outlier (lower right graph). C:
Image data with a structural outlier (lower right image).}
\end{figure}

The framework we propose generalizes an approach for outlier detection
in functional data developed recently
(\protect\hyperlink{ref-herrmann2021geometric}{Herrmann \& Scheipl,
2021}). Since the approach exploits the metric structure of a functional
data set, it is straightforward to generalize it to other data types,
both from a theoretical as well as a practical perspective.
Theoretically, the observation space needs to be a metric space, i.e.~it
needs to be equipped with a metric. Practically, there only needs to be
a suitable distance measure to compute pairwise distances between
observations. Two assumptions are fundamental for the framework. First
of all, the \emph{manifold assumption} that observed high-dimensional
data lie on or close to a (low-dimensional) manifold. Note that
functional data typically contain a lot of structure, and it is often
reasonable to assume that only a few modes of variation suffice to
describe most of the information contained in the data, i.e., such
functional data often have low intrinsic dimension, at least
approximately, see Figure \ref{fig:outtypes} for a simple synthetic
example. Similar remarks hold for other data types such as image data
(\protect\hyperlink{ref-lee2007nonlinear}{Lee \& Verleysen, 2007};
\protect\hyperlink{ref-ma2011manifold}{Ma \& Fu, 2011}). Secondly, it is
assumed that outliers are either \emph{structural outliers} -- or in the
terminology of Zimek and Filzmoser
(\protect\hyperlink{ref-zimek2018there}{2018, p. 10}) ``real outliers''
stemming from a different data generating process than the bulk of the
data -- or \emph{distributional} outliers, observations that are
structurally similar to the main data but still appear outlying in some
sense. We make these notions mathematically precise in the remainder of
this section based on the exposition in Herrmann \& Scheipl
(\protect\hyperlink{ref-herrmann2021geometric}{2021}) before we
demonstrate the practical relevance of the framework in section
\ref{sec:exps} and summarize its general conceptual implications in
section \ref{sec:discussion}.\\
Given a high-dimensional observation space \(\hdspace\) of dimension
\(\obsdim\), a \(d\)-dimensional parameter space
\(\pspace \subset \mathbb{R}^d\), such that the elements
\(\theta_i \in \pspace\) are realizations of the probability
distribution \(P\) over the domain \(\mathbb{R}^d\), i.e.,
\(\theta_i \sim P\), and given an embedding space
\(\embedspace \subset \mathbb{R}^{d'}\), define the mappings \(\phi\)
and \(e\) so that
\[\Theta \stackrel{\phi}{\to} \mathcal{\mani_{\hdspace}} \stackrel{e}{\to} \mathcal{Y},\]
with \(\mani_{\hdspace} \subset \hdspace\) a manifold in the observation
space. The structure of \(\mani_{\hdspace}\) is determined by the
structure and dimensionality of \(\pspace\), \(P\), and the map
\(\phi\), which is isometric for the appropriate metrics in \(\pspace\)
and \(\hdspace\). Conceptually, the low-dimensional parameter space
\(\Theta\) represents the modes of variation of the data and the mapping
\(\phi\) represents the data generating process that yields
high-dimensional data \(x_i = \phi(\theta_i) \in \mani_{\hdspace}\)
characterized by these modes of variation. We assume that
low-dimensional representations of the observed data in the embedding
space \(\embedspace\), which capture as much of the metric structure of
\(\mani_{\hdspace}\) as possible, can be learned from the observed data.
A successful embedding \(e\) then also recovers as much of the structure
of the parameter space \(\pspace\) as possible in the low dimensional
representations \(y_i = e(x_i) \in \embedspace\).

In our framework, distributional outliers are defined w.r.t. minimum
volume sets (\protect\hyperlink{ref-polonik1997minimum}{Polonik, 1997})
of \(P\) in this parameter space \(\pspace\):

\noindent \textbf{Definition 1:} \emph{Minimum volume set}\\
Given a probability distribution \(P\) over (a subset of)
\(\mathbb{R}^d\), a minimum volume set \(\Omega^*_{\alpha}\) is a set
that minimizes the quantile function
\(V(\alpha) = \inf_{C \in \mathcal{C}}\{\text{Leb}(C): P(C) \geq \alpha\}, 0 < \alpha < 1\)\}
for i.i.d. random variables in \(\mathbb{R}^{d}\) with distribution
\(P\), \(\mathcal{C}\) a class of measurable subsets in
\(\mathbb{R}^{d}\) and Lebesgue measure \(\text{Leb}\).\\
So \(\Omega^*_{\alpha, P}\) is the smallest region containing a
probability mass of at least \(\alpha\). We can now define structural
outliers and distributional outliers as follows:

\noindent \textbf{Definition 2:} \emph{Structural and distributional
outlier}\\
Define \(\mani_{\Theta, \phi}\) as the codomain of \(\phi\) applied to
\(\Theta\).\\
Define two such manifolds \(\Man = \mani_{\Than, \phan}\) and
\(\Min = \mani_{\Thin, \phin}\) and a data set
\(X \subset \Man \cup \Min\).\\
W.l.o.g., let
\(r = \frac{\vert \{x_i: x_i \in \Man \land x_i \notin \Min\} \vert}{\vert \{x_i: x_i \in \Min\} \vert} \lll 1\)
be the \emph{structural outlier ratio}, i.e.~most observations are
assumed to stem from \(\Min\). Then an observation \(x_i \in X\) is

\begin{itemize}
\tightlist
\item
  a \emph{structural outlier} if \(x_i \in \Man\) and
  \(x_i \notin \Min\) and
\item
  a \emph{distributional outlier} if \(x_i \in \Min\) and
  \(\theta_i \notin \Omega^*_{\alpha}\), where \(\Omega^*_{\alpha}\) is
  defined by the density of the distribution generating \(\Than\).
\end{itemize}

Figure \ref{fig:outtypes} shows examples of three data types with
structural outliers (in red) and some distributional outliers for the
functional data example. Since distributional outliers are structurally
similar to inliers, they are hard to detect visually for graph and image
data, as doing so requires a lot of ``normal'' data to reference against
and we can only display a few example observations here. \color{black} As
outlined, \color{black}this is one reason why we \color{black}again
\color{black} use functional data for our exposition in the following.

Summarizing the framework's crucial aspects in less technical terms, we
assume that the bulk of the observations comes from a single ``common''
process, which generates observations in some \color{black} subset
\color{black} \(\Min\), while some data might come from an ``anomalous''
process, which defines structurally distinct observations in a different
\color{black} subset \color{black} \(\Man\). This follows standard notions
in outlier detection which often assume (at least) two different
data-generating processes (\protect\hyperlink{ref-dai2020functional}{Dai
et al., 2020}; \protect\hyperlink{ref-zimek2018there}{Zimek \&
Filzmoser, 2018}). Note that this does not imply that structural
outliers are in any way similar to each other: \(\Pan\) could be very
widely dispersed or arise from a mixture or several different
distributions and/or \(\Man\) could consist of several unconnected
components representing various kinds of structural abnormality. The
only crucial aspect is that the process from which \emph{most} of the
observations emerge yields structurally similar data. We consider
settings with a structural outlier ratio \(r \in [0, 0.1]\) to be
suitable for outlier detection. The proportion of \emph{distributional}
outliers on \(\Min\), in contrast, depends only on the \(\alpha\)-level
for \(\Omega^*_{\alpha, \Pin}\). Practically speaking, neither prior
knowledge about these manifolds nor specific assumptions about
structural differences are necessary for our approach. The key points
are that (1) structural outliers are not on the main data manifold
\(\Min\), (2) distributional outliers are at the edges of \(\Min\), and
(3) these properties are preserved in the embedding vectors as long as
the embedding is based on an appropriate notion of distance in
\(\hdspace\).

\section{Experiments}\label{sec:exps}

This section lays out practical implications of the framework through
experiments on several different data types, via a comprehensive
qualitative and visual analysis of six examples. In addition, we provide
quantitative results for six labeled data sets.

\subsection{Methods}\label{sec:exps:meths}

The focus of our experiments is to evaluate a general framework for
outlier detection, which is motivated by geometrical considerations.
With these experiments, we support the claim that the perspective
induced by the framework lets us visualize, detect, and analyze outliers
in a principled and canonical way. For this demonstration, we chose
Multidimensional Scaling (MDS)
(\protect\hyperlink{ref-cox2008multidimensional}{Cox \& Cox, 2008}) as
our embedding method and Local Outlier Factors (LOF)
(\protect\hyperlink{ref-breunig2000lof}{Breunig et al., 2000}) as our
outlier scoring method. Note that the experiments are not intended to
draw conclusions about the superiority of these specific methods and
other combinations of methods may be as suitable or even superior for
the purpose (see for example results for Isomap in Herrmann \& Scheipl
(\protect\hyperlink{ref-herrmann2021geometric}{2021})).\\
However, more sophisticated embedding methods than MDS require tuning
over multiple hyperparameters, whereas MDS has only one -- the embedding
dimension. Moreover, an advantage of MDS over other embedding methods is
that it aims for isometric embeddings, i.e., tries to preserve all
pairwise distances as closely as possible, which is crucial in
particular to reflect structural outlyingness. In fact, Torgerson
Multidimensional Scaling (tMDS, i.e., MDS based on \(L_2\) distance) --
that is: a simple linear embedding equivalent to standard PCA scores --
seems to uncover many outlier structures sufficiently well in many data
settings despite its simplicity. For similar reasons, we chose to use
LOF as an outlier scoring method. This method also has a single
hyperparameter, \(minPts\), the number of nearest neighbors used to
define a point's (local) neighborhood, which we denote as \(k\) in the
following. Moreover, in contrast to many other outlier scoring methods
such as one-class support vector machines
(\protect\hyperlink{ref-munoz2004one}{Muñoz \& Moguerza, 2004}) which
require low-dimensional tabular data as input (i.e.~which can only be
applied to complex data types indirectly by using embedding vectors as
feature inputs), LOF can also be applied to high-dimensional and
non-tabular data directly as it only requires a distance matrix as
input. Experiments on functional data have shown that LOF applied
directly to a distance matrix of functional data and LOF applied to the
corresponding embedding vectors yield consistent results
(\protect\hyperlink{ref-herrmann2021geometric}{Herrmann \& Scheipl,
2021}).\\
Note, however, that beyond the ability to apply outlier scoring methods
to low-dimensional embedding vectors of high-dimensional and/or
non-tabular data, such embeddings provide additional practical value: In
particular, scalar scores or ranks as provided by outlier scoring
methods are not able to reflect differences between distributional and
structural outliers whereas such differences become accessible and
interpretable in visualizations of these embeddings.\\
This also points to a major caveat of the quantitative (in contrast to
the qualitative) experiments, in which we use ROC-AUC to evaluate the
accuracy of outlier ranks obtained with LOF with respect to the
``outlier'' structure defined by the different classes of labeled data.
Setting one class as \(\Min\) and contaminating this ``normal'' class
with observations from other classes, which are assumed to be
structurally different and thus form \(\Man\), we obtain data sets
\(X \subset \Min \cup \Man\). Although this is a widely used approach
(\protect\hyperlink{ref-campos2016evaluation}{Campos et al., 2016};
\protect\hyperlink{ref-goldstein2016comparative}{Goldstein \& Uchida,
2016}; \protect\hyperlink{ref-pang2018learning}{Pang et al., 2018}),
such an evaluation only considers outliers as defined by the class
labels and poor AUC values may not necessarily imply poor performance if
there are observations from the ``normal'' data class which are
(distributionally or structurally) more outlying (and thus obtain higher
scores) than some of the ``labeled'' outliers, see also Campos et al.
(\protect\hyperlink{ref-campos2016evaluation}{2016}). This is why we do
not merely report potentially problematic quantitative measures (Section
\ref{sec:exps:quant}), and instead put more emphasis on qualitative
experiments that are much closer to the way we would recommend using
these methods in practical applications.

\subsection{Qualitative assessment}\label{sec:exps:qual}

In this section, we provide extensive qualitative analyses to
demonstrate the practical relevance of the framework. First, we
demonstrate that the distinction between structural and distributional
outliers is preserved in embeddings using two simulated functional data
sets. Secondly, using two real-world data sets -- a functional and an
image data set -- we show that the approach can be applied flexibly to
different data structures. Thirdly, we illustrate the general
applicability to more general data types based on synthetic graph data
and real-world curves data. In the following, all LOF results are
obtained using \(k = 0.75n\), where \(n\) is the number of observations.

\subsubsection{Demonstrating the framework's practical implications on idealized synthetic data}

\begin{figure}
\centering
\includegraphics{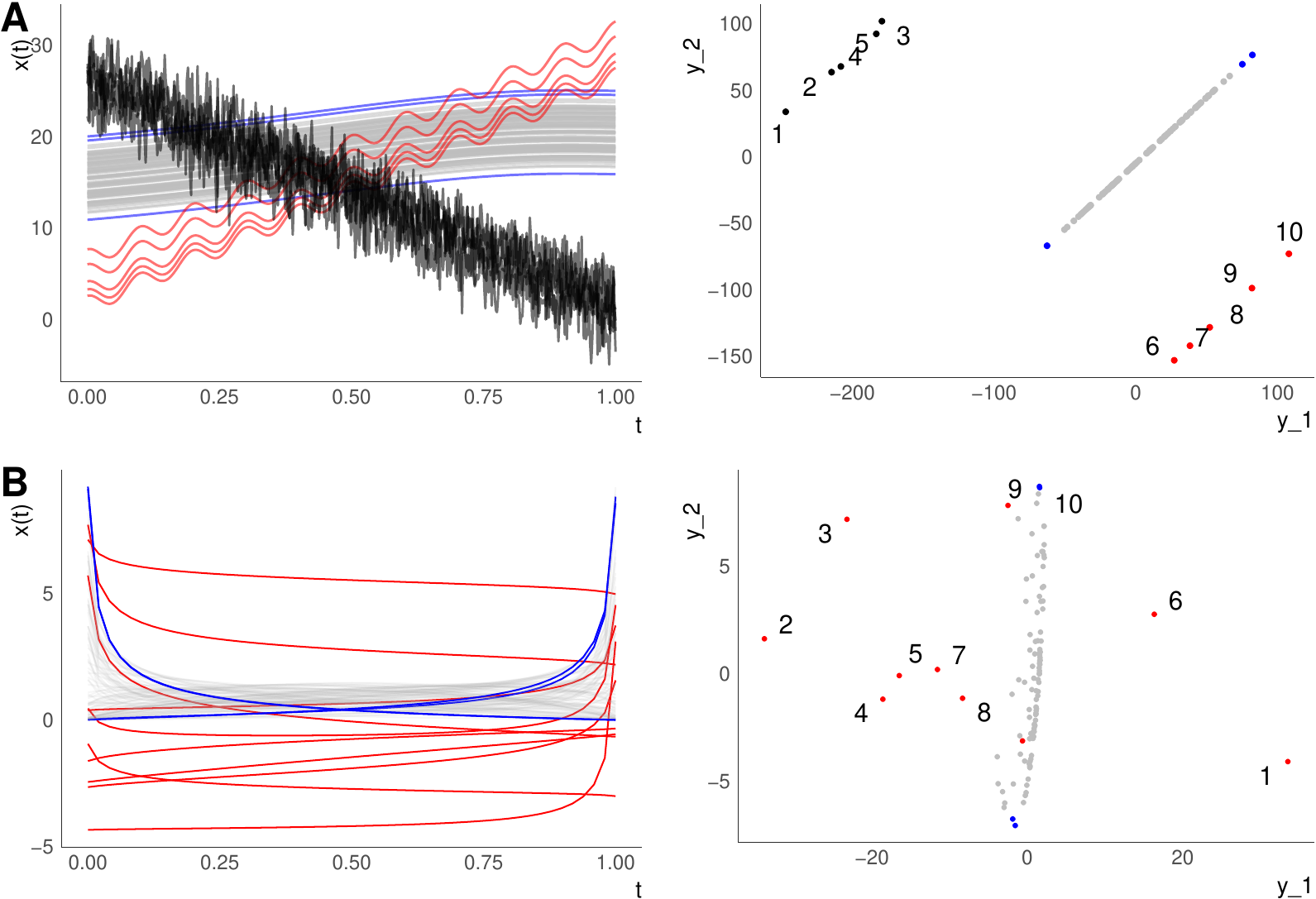}
\caption{\label{fig:exps-sim} Simulated functional data and their
2\(\vizdim\) embeddings. Numbered labels are ascending LOF score ranks
of the outliers (\(k = 0.75n\)).}
\end{figure}

Figure \ref{fig:exps-sim} shows two simulated functional data sets (A \&
B, left) and their 2\(\vizdim\) PCA/tMDS embeddings (A \& B, right). One
can observe that data set A is an example with structural outliers in
terms of shape and slope. This is an extended version of an example by
Hernández \& Muñoz (\protect\hyperlink{ref-hernandez2106kernel}{2016})
and, following their notation, the two manifolds can be defined as
\(\Man = \{x(t) \vert x(t) = b + 0.05t + \cos(20\pi t), b \in \mathbb R\} \cup \{x(t) \vert x(t) = (c - 0.05t) + e_t, c, e_t \in \mathbb R\}\)
and
\(\Min = \{x(t) \vert x(t) = a + 0.01t + \sin(\pi t^2), a \in \mathbb R\}\)
with \(t \in [0, 1]\) and \(a \sim N(\mu = 15, \sigma = 4)\),
\(b \sim N(\mu = 5, \sigma = 3)\), \(c \sim N(\mu = 25, \sigma = 3)\),
and \(e_t \sim N(\mu = 0, \sigma = 4)\). Note that the structural
outliers are not all similar to each other in shape or slope, which is
reflected in \(\Min\) being a union of two structurally different
manifolds.\\
In contrast, data set B of various (vertically shifted)
Beta-distribution densities is an example where distributional
outlyingness is defined by phase -- i.e.~horizontal -- variation and
structural outlyingness by vertical shifts. The respective manifolds are
defined as
\(\Man = \{x(t) \vert x(t) = b + B(t, \alpha, \beta), (b, \alpha, \beta)' \in \mathbb R \times \mathbb R^+ \times \mathbb R^+\}\)
and
\(\Min = \{x(t) \vert x(t) = B(t, \alpha, \beta), (\alpha, \beta)' \in \mathbb R^+ \times \mathbb R^+\}\)
with \(t \in [0, 1]\), \(\alpha, \beta \sim U[0.1, 2]\),
\(b \sim U[-5, 5]\) and \(B\) the density of the beta distribution. For
both, we generate 100 ``normal'' observations from \(\Min\) and 10
structural outliers from \(\Man\), with \(\obsdim = 500\) evaluation
points in the first and \(\obsdim = 50\) evaluation points in the latter
example.\\
Structural outliers are clearly separated from observations on \(\Min\)
in both cases and appear as outlying in the 2\(\vizdim\) embeddings.
Moreover, we see that distributional outliers are embedded at the
periphery of \(\Min\). Numbers in the figures are ascending LOF score
ranks of the outliers. Note that \(\Min \subset \Man\) in data set B.
Nevertheless, most structurally outlying observations from \(\Man\) are
clearly separated in the embedding. Two structural outliers are in or
very close to \(\Min \cup \Man\) and thus appear in the main bulk of the
data. The LOF scores also reflect this, as one of the distributional
outliers is ranked as even more outlying.

Summarizing, we see that in these simulated situations, practically
relevant outlier sub-structure -- deviations in terms of functional
shape, slope, or vertical shifts -- are represented accurately by
low-dimensional embeddings learned from the observed high-dimensional
data. \color{black} In particular, structural outliers do not need to be
similar to each other as Example A demonstrates. Also, note that Example
B illustrates as a by-product that there can be situations where the
approach yields meaningful results even though the two manifold are not
completely disjoint. However, this does not necessarily hold in general.
See Souvenir \& Pless
(\protect\hyperlink{ref-souvenirManifoldClustering2005}{2005}) for an
approach to disentangle intersecting manifolds. \color{black} Moreover,
we see that situations where distributional outliers appear ``more''
outlying than structural outliers are captured as well. Note that this
is a crucial aspect. Although this aspect is quantified correctly by an
outlier scoring method such as LOF, the two outlier types can be
distinguished only if visualizations, as provided by embedding methods,
are considered. Consider that evaluation of unsupervised outlier
detection is often performed using a labeled data set, setting
observations from one class as inliers and sampling observations from
another class as outliers, and then computing binary classification
performance measures such as the AUC
(\protect\hyperlink{ref-campos2016evaluation}{Campos et al., 2016};
\protect\hyperlink{ref-goldstein2016comparative}{Goldstein \& Uchida,
2016}; \protect\hyperlink{ref-pang2018learning}{Pang et al., 2018}).
Different class labels do not guarantee that the classes do not overlap,
i.e., that the respective manifolds are disjoint in \(\hdspace\), nor
that there are no distributional outliers appearing more outlying than
structural outliers. Thus, there may be distributional outliers among
the inliers which are scored as more outlying than structural outliers
(see data set B) and a purely quantitative assessment is likely to
mislead. Being able to create faithful visualizations of such more
complex outlier structures for high-dimensional data is a crucial
benefit of the proposed approach.

\subsubsection{Demonstrating flexibility on real functional and image data}

\begin{figure}
\centering
\includegraphics{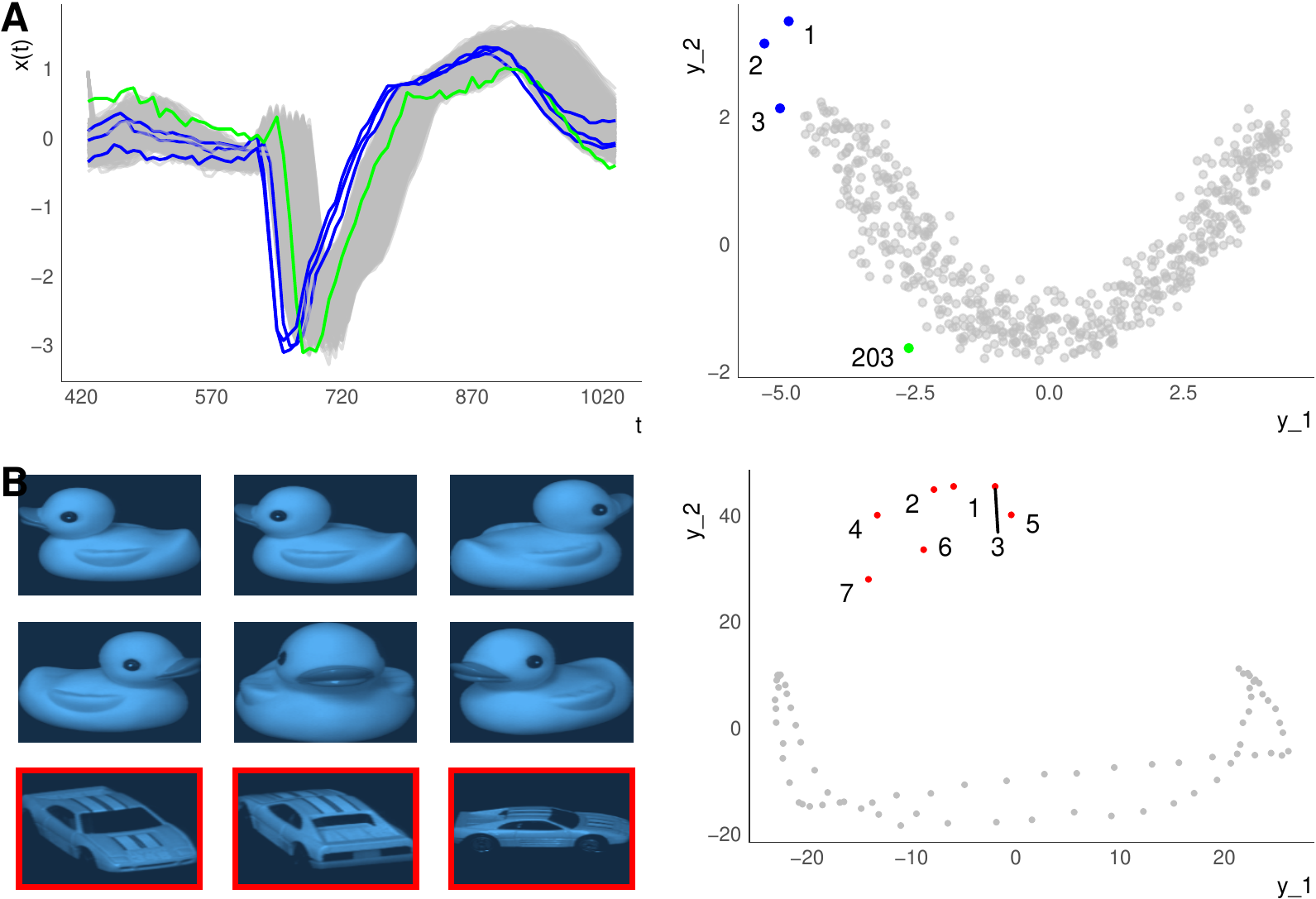}
\caption{\label{fig:fda-image-real} Real functional and image data and
their 2\(\vizdim\) tMDS embeddings. Numbered labels are ascending LOF
score ranks of the outliers (\(k = 0.75n\)).}
\end{figure}

Of course, real-world data settings are usually more complicated than
our simulated examples. First of all, real data are much more difficult
to assess since the underlying manifolds are usually not directly
accessible, so it is impossible to define the exact structure of the
data manifolds like in the simulated examples. In addition, some data
sets may not contain any clear \textit{structural} outliers, while
others may not contain any clear \textit{distributional} outliers, or
both. A crucial aspect of the approach is that, although it is based on
a highly abstract conceptualization involving unobservables like the
parameter space \(\Theta\) and its probability measure \(P\), it is not
at all necessary to come up with any such formalization of the data
generating process to put the approach into practice and obtain
meaningful results, as will be demonstrated in the following. \\
Consider Figure \ref{fig:fda-image-real}, which shows a real functional
data set of 591 ECG measurements (\protect\hyperlink{ref-dau2019ucr}{Dau
et al., 2019};
\protect\hyperlink{ref-goldberger2000physiobank}{Goldberger et al.,
2000}) with 82 evaluation points per function, i.e.~a \(\obsdim = 82\)
dimensional data set (A), and a sample of the COIL20 data
(\protect\hyperlink{ref-coil20}{Nane et al., 1996}) (B). It is
impossible to define the exact structure of the ECG data manifold.
However, the visualizations of the functions on the left-hand side
suggest that there are no observations with clear structural differences
in functional form: none of the curves are clearly shifted away from the
bulk of the data, nor are there any curves with isolated peaks, or
observations with clearly different shapes. In accordance with this
observation, there is also no clearly separable structure in the
embedding. However, observations that appear in low-density regions of
the embedding can be regarded as distributional outliers in terms of
horizontal shift, i.e., phase variation, like the three observations
with the earliest minima colored in blue. This is also reflected in the
scoring of the embeddings, as the observations with the lowest LOF ranks
are clear distributional outliers in function space. However, the
embedding provides much more complete information in this example than
LOF ranks and the functional visualization alone. For example, they also
pinpoint a \textit{vertical shift} outlier in the first and last thirds
of the domain (green curve, which would be hard to detect based on its
functional representation alone). This apparently represents a second
``dimension'' of distributional outlyingness.\\
The COIL20 data (\protect\hyperlink{ref-coil20}{Nane et al., 1996})
consists of 1440 pictures (\(128 \times 128\), grayscale) of 20
different objects. The \(72\) pictures in each class depict one and the
same object at different rotation angles with a picture taken at every
\(5\)° within \([0\)°\(, 355\)°\(]\). We use all \(72\) pictures of a
rubber duck to represent observations from \(\Min\) and randomly sample
\(7\) observations (i.e.~\(r \approx 0.1\)) from the \(72\) pictures of
a toy car as structural outliers from \(\Man\). We compute \(L_2\)
distances of the vectorized pixel intensities
(\(\obsdim = 128^2 = 16384\)). Figure \ref{fig:fda-image-real} B, left
column, shows a sample of \(6\) inlier and \(3\) structural outlier
pictures, the right column shows embeddings of all \(79\) images. Since
the inlier data are images of a rotated object, \(\Min\) is the image of
a one-dimensional closed and circular parameter space defining the
rotation angle (c.f. \protect\hyperlink{ref-ma2011manifold}{Ma \& Fu,
2011}), i.e., other than in the ECG example substantial considerations
yield at least some knowledge about the specific structure of the data
manifold(s) in this case.

The 2\(\vizdim\) embedding reflects the expected structure of our COIL20
subset very well, with clear separation of the \(7\) pictures of the toy
car as structural outliers. In addition, the embedding of \(\Min\)
indeed yields a closed, but not quite a circular loop, as does the
embedding of the 7 rotated images from \(\Man\). The corresponding 3D
embedding (not shown) reveals that the embeddings of the inliers lie on
a rough circle folded over itself. In summary, in the ECG example there
seem to be no clearly separable, structurally different outliers that
could be detected with tMDS, but only distributional outliers, whereas
in the COIL data there are clearly separate structural outliers, but no
distributionally outlying observations. These two examples with very
different intrinsic structures (single connected manifold with
distributional outliers versus disconnected manifolds without clear
distributional outliers) illustrate that it is not necessary to have
explicit prior knowledge about the data generating process or its
outlier characteristics for the approach to work and that it is able to
handle different data manifold structures flexibly and successfully.

\subsubsection{Demonstrating generalizability on graph and curve data}

\begin{figure}
\centering
\includegraphics{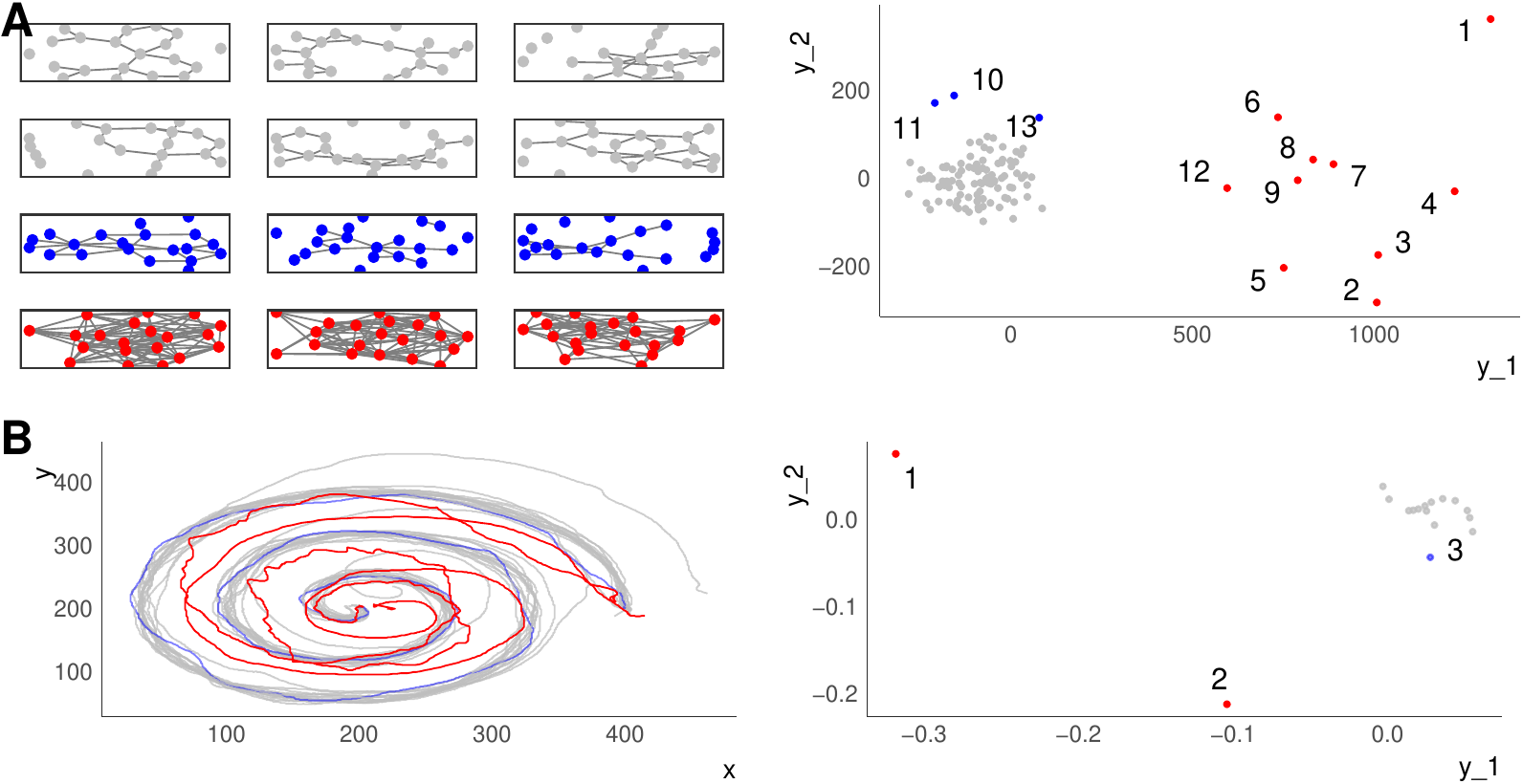}
\caption{\label{fig:further-qual-exp} Curve and graph data as further
examples to demonstrate the flexibility and general applicability of the
approach, and their 2\(\vizdim\) MDS embeddings based on Frobenius
(graphs) and Elastic shape distances (curves). Numbered labels are
ascending LOF score ranks of the outliers (\(k = 0.75n\)).}
\end{figure}

Note that the COIL example illustrates that the framework also works in
image data and that a fairly simplistic approach of computing \(L_2\)
distances between vectorized pixel intensities yields very reasonable
results in this example. The framework is, however, not at all
restricted to these two data types nor such a simple distance metric.
Recall that the approach can be applied to any data type whatsoever as
long as a suitable distance metric is available. Beyond 1\(\vizdim\)
functional and image data, the framework can also be extended to more
general and complex data types, for example, graphs or 2\(\vizdim\)
curves as depicted in Figures \ref{fig:further-qual-exp}. We use more
specialized distance measures to show that good results can also be
obtained on such data.\\
We simulate two structurally different classes of Erd\H{o}s-Rényi graphs
with 20 vertices (see Fig. \ref{fig:further-qual-exp} A). This
structural difference results from different edge probabilities \(p_v\)
that two given vertices of the graph are connected, setting
\(p_{v} = 0.1\) for \(\Min\) and \(p_{v} = 0.4\) for \(\Man\). We
randomly sample \(100\) observations from \(\Min\) and \(10\) from
\(\Man\), i.e.~\(r = 0.1\), and obtain a pairwise distance matrix by
computing the Frobenius distances between the graph Laplacians.

The curves data (Fig. \ref{fig:further-qual-exp} B) consists of spiral
curve drawings from an Archimedes spiral-drawing test that is used to
diagnose patients with Parkinson's disease
(\protect\hyperlink{ref-alty2017use}{Alty et al., 2017};
\protect\hyperlink{ref-steyer2021elastic}{Steyer et al., 2021}). Taking
data from the dynamic version of the test
(\protect\hyperlink{ref-isenkul2014improved}{Isenkul et al., 2014}), we
use 15 curves drawn by healthy controls not suffering from Parkinson's
disease and two curves drawn by Parkinson patients to represent
potential structural outliers, where each curve is evaluated on 200
points. Previous investigations have shown that an elastic shape
distance is better suited than \(L_2\) distances to discriminate between
the two groups (\protect\hyperlink{ref-steyer2021elastic}{Steyer et al.,
2021}).\\
So, in contrast to the previous examples, we use more specialized
distance measures to capture the relevant structures in these settings.
This illustrates that the approach is not only flexible with respect to
the actual structure present in a given data set as demonstrated in the
previous section but that it is also very generally applicable to a
variety of data types. The approach can be used for any kind of data
simply by defining an appropriate (data-specific) distance measure. In
both, the embeddings of the graphs, as well as the embeddings of the
curves, structurally different observations (in red) are clearly
separated from the observations on \(\Min\). This is also reflected by
their LOF scores. Moreover, in both settings, there are observations
from \(\Min\) (in blue) which appear in peripheral, sparser regions of
the ``normal'' data and thus can be considered distributional outliers.
Note that it is not always immediately obvious on the level of the
original data why observations appear distributionally outlying. For
example, in the graph data, note that other than in previous examples
(e.g.~Fig. \ref{fig:exps-sim} A) comparing them to a few inliers does
not reveal a striking difference at first (in contrast to the structural
outliers!): Figure \ref{fig:further-qual-exp} A, left column, shows six
inlier graphs in the 1st and 2nd row, the three distributional outlier
graphs in the 3rd row, and three structural outlier graphs in the 4th
row.\\
Nevertheless, the embedding vectors and their LOF ranks indicate that
the distributionally outlying observations have obtained some specific
characteristics setting them apart from most inlying observations. For
example, further analysis reveals that the graph with LOF rank 11
contains the node with maximum connectedness of all nodes in all inlier
graphs. Its degree is 8 (i.e., it is directly connected to 8 other
nodes), while the average of the maximum degree in the graphs on
\(\Min\) is just 4.39. In contrast, the graph with LOF rank 13 contains
8 isolated nodes of degree 0, while the average number of nodes with
degree 0 is only 2.47 on \(\Min\). The respective values of the graph
with LOF rank 10 are above the upper quartile for both of these metrics,
with 4 unconnected nodes and a maximally connected node with degree 6.

\hypertarget{quantitative-assessment}{%
\subsection{\texorpdfstring{Quantitative assessment
\label{sec:exps:quant}}{Quantitative assessment }}\label{quantitative-assessment}}

In order to provide less subjective experimental results, we assess the
approach quantitatively, using labeled data with at least two classes.
For each data set, we consider four outlier ratios
\(r \in \{0.01, 0.025, 0.5, 0.1\}\). Setting one class as \(\Min\), with
\(n_{in} = \vert \Min \vert\), and contaminating this ``normal'' class
with \(n_{out} = r \cdot n_{in}\) ``structural'' outliers from other
classes, which form \(\Man\), we obtain data sets
\(X \subset \Min \cup \Man\) with \(n = n_{in} + n_{out}\). For each
setting, we repeat the contamination process 50 times, sampling outliers
at random from \(\Man\). Based on outlier ranks computed with LOF, we
use ROC-AUC as a performance measure and report the mean AUCs over the
50 replications for each combination of settings. Note that we only use
the labels of the ``structural'' outliers for computing this performance
measure, not for the unsupervised learning of the embeddings themselves.
For all data sets considered in this section, plots of typical
embeddings for \(r = 0.05\) can be found in Figures \ref{fig:app} and
\ref{fig:app-tab} in appendix \ref{sec:app-1}. We consider three
additional functional data sets for this experiment: \textit{dodgers}
(\protect\hyperlink{ref-dau2019ucr}{Dau et al., 2019}), a set of times
series of daily traffic close to Dodgers Stadium, with days on weekends
forming \(\Man\) and weekdays forming \(\Min\); \textit{phoneme}
(\protect\hyperlink{ref-febrero2012fdausc}{Febrero-Bande \& Oviedo de la
Fuente, 2012}), discretized log-periodograms of five different phonemes,
with phoneme ``dcl'' forming \(\Man\) and phonemes ``sh'', ``iy'',
``aa'', and ``ao'' forming \(\Min\); \textit{starlight}
(\protect\hyperlink{ref-dau2019ucr}{Dau et al., 2019};
\protect\hyperlink{ref-rebbapragada2009finding}{Rebbapragada et al.,
2009}), phase-aligned light curves of Eclipsing Binary, Cepheid, and RR
Lyrae stars, the first forming \(\Man\) and the latter two forming
\(\Min\). All results are based on simple, linear tMDS/PCA embeddings
with the LOF algorithm applied to the resulting 2\(\vizdim\) embedding
vectors.\\
\color{black} In addition, we consider three tabular data sets, two real
and one simulated. This includes the well-known Iris data
(\protect\hyperlink{ref-anderson1935irises}{Anderson, 1935};
\protect\hyperlink{ref-fisher1936use}{Fisher, 1936}) where class
\emph{Setosa} forms \(\Min\) and the other two classes \(\Man\).
Moreover, we use the Wiscon Breast Cancer (wbc) data
(\protect\hyperlink{ref-streetNuclearFeatureExtraction1993}{Street et
al., 1993}) as provided by the UCI Machine Learning repository
(\protect\hyperlink{ref-Dua2019}{Dua \& Graff, 2017}). This tabular data
set comprises 30 features containing information about the cell nuclei
of breast tissue and has been used by Goldstein \& Uchida
(\protect\hyperlink{ref-goldstein2016comparative}{2016}) for outlier
detection before. Following their approach, the healthy patients form
\(\Min\) and patients with malignant status form \(\Man\). Yet, other
than Goldstein \& Uchida
(\protect\hyperlink{ref-goldstein2016comparative}{2016}), we do not fix
outliers to the first 10 observations from the latter class but -- as
outlined -- repeatedly sample outliers at random from \(\Man\). Finally,
we include a simple simulated example where
\(\Min = \{x : x \sim \mathcal{N}_{1000}(\mathbf{0}, \Sigma)\}\) and
\(\Man = \{x : x \sim \mathcal{N}_{1000}(\mathbf{1}, \Sigma)\}\),
\(\Sigma = \text{diag}(\mathbf{1})\). That is, 1000-dimensional data
with observations sampled from two multivariate normal distributions
where the class difference stems from the difference in the mean
vectors, \(\mathbf{0}\) for \(\Min\) and \(\mathbf{1}\) for \(\Man\).\\
\color{black} The results depicted in Table \ref{tab:real-quant} show
that outlier detection does not need to be specifically challenging in
nominally high-dimensional data. In each of the data sets, which have
very different numbers of observations and numbers of dimensions, high
ROC-AUC \(\geq 0.95\) can be achieved for all considered outlier ratios
\(r\). This indicates that most of the observations from \(\Man\) indeed
appear to be outlying in the embedding space and thus obtain high LOF
scores. Furthermore, as in the qualitative analysis, a global setting of
\(k = 0.75n\) seems to be a reasonable default for the LOF algorithm.
Only for \(r = 0.01, 0.025\) in the starlight data, we see a large
improvement (AUC \(= 1.00\)) with \(k = 0.1n\). For small \(r < 0.1\),
in all other settings the achieved ROC-AUC is very robust against
changes in this tuning parameter.

\begin{table}
\scriptsize
\caption{\label{tab:real-quant}Mean ROC-AUC values over 50 replications based on the ranks as assigned by LOF. Each data set consists of $n$ observations, $n_{in}$ from $\Min$ and $n_{out} = n_{in} \cdot r$ from $\Man$. $\Man$ and $\Min$ are defined by classes of the original labeled data sets. $D$ is the dimensionality of a data set (i.e, evaluations per function for functional data) and $k$ the number of nearest neighbors used in the LOF algorithm. A: Functional data. B: Tabular data.}
\centering
\resizebox{\textwidth}{!}{\begin{tabular}{@{}lcccccrccccrcccc@{}} \toprule
\textbf{A} & & \multicolumn{4}{c}{dodgers} & & \multicolumn{4}{c}{phoneme} & & \multicolumn{4}{c}{starlight} \\
 & & \multicolumn{4}{c}{$n_{in} = 97$, $\obsdim = 289$} & & \multicolumn{4}{c}{$n_{in} = 400$, $\obsdim = 150$} & & \multicolumn{4}{c}{$n_{in} = 6656$, $\obsdim = 1025$} \\
\cmidrule{3-6} \cmidrule{8-11} \cmidrule{13-16}
 & $k$       & $0.01n$ & $0.1n$ & $0.75n$ & $0.9n$ & & $0.01n$ & $0.1n$ & $0.75n$ & $0.9n$ & & $0.01n$ & $0.1n$ & $0.75n$ & $0.9n$ \\  \midrule
$r: 1.0\%$ & & 0.78 & 0.98 & 0.96 & 0.96 & & 0.78 & 1.00 & 0.99 & 0.99 & & 0.96 & 1.00 & 0.69 & 0.78 \\
$r: 2.5\%$ & & 0.62 & 0.97 & 0.96 & 0.96 & & 0.54 & 1.00 & 0.99 & 0.99 & & 0.55 & 1.00 & 0.88 & 0.88 \\
$r: 5.0\%$ & & 0.59 & 0.97 & 0.96 & 0.96 & & 0.56 & 0.99 & 0.99 & 0.99 & & 0.53 & 1.00 & 0.92 & 0.92 \\
$r: 10\%$  & & 0.54 & 0.84 & 0.97 & 0.96 & & 0.57 & 0.75 & 0.99 & 0.99 & & 0.56 & 0.98 & 0.95 & 0.87\\
\end{tabular}}
\color{black}
\resizebox{\textwidth}{!}{\begin{tabular}{@{}lcccccrccccrcccc@{}} \toprule
\textbf{B} & & \multicolumn{4}{c}{iris} & & \multicolumn{4}{c}{wisconsin breast cancer} & & \multicolumn{4}{c}{simulated data} \\
 & & \multicolumn{4}{c}{$n_{in} = 50$, $\obsdim = 4$} & & \multicolumn{4}{c}{$n_{in} = 357$, $\obsdim = 30$} & & \multicolumn{4}{c}{$n_{in} = 750$, $\obsdim = 1000$} \\
\cmidrule{3-6} \cmidrule{8-11} \cmidrule{13-16}
 & $k$       & $0.01n$ & $0.1n$ & $0.75n$ & $0.9n$ & & $0.01n$ & $0.1n$ & $0.75n$ & $0.9n$ & & $0.01n$ & $0.1n$ & $0.75n$ & $0.9n$ \\  \midrule
$r: 1.0\%$ & & 1.00 & 1.00 & 1.00 & 1.00 & & 0.76 & 0.96 & 0.94 & 0.89 & & 0.66 & 1.00 & 1.00 & 1.00 \\
$r: 2.5\%$ & & 0.71 & 1.00 & 1.00 & 1.00 & & 0.64 & 0.97 & 0.95 & 0.92 & & 0.56 & 1.00 & 1.00 & 1.00 \\
$r: 5.0\%$ & & 0.52 & 1.00 & 1.00 & 1.00 & & 0.60 & 0.97 & 0.94 & 0.91 & & 0.58 & 1.00 & 1.00 & 1.00 \\
$r: 10\%$  & & 0.61 & 0.69 & 1.00 & 1.00 & & 0.58 & 0.96 & 0.94 & 0.92 & & 0.56 & 1.00 & 1.00 & 1.00 \\
\bottomrule
\end{tabular}}
\color{black}
\end{table}

\hypertarget{sec:discussion}{%
\section{Discussion}\label{sec:discussion}}

\hypertarget{sec:discussion:implications}{%
\subsection{Summary}\label{sec:discussion:implications}}

We propose a geometrically motivated framework for outlier detection,
which exploits the metric structure of a (possibly high-dimensional)
data set and provides a mathematically precise distinction between
\emph{distributional} outliers and \emph{structural} outliers.
Experiments show that the outlier structure of high-dimensional and
non-tabular data can be detected, visualized, and quantified using
established manifold learning methods and standard outlier scoring. The
decisive advantage of our framework from a theoretical perspective is
that the resulting embeddings make subtle but important properties of
outlier structure explicit and -- even more importantly -- that these
properties are made accessible based on visualizations of the
embeddings. From a more practical perspective, our proposal requires no
prior knowledge nor any specific assumptions about the actual data
structure in order to work, an important aspect since data generating
processes are usually inaccessible. This is highly relevant in practice,
in particular since a well-established, computationally cheap
combination of widely used and fairly simple methods like (t)MDS and LOF
proved to be a strong baseline that yields fairly reliable results
without the need for tuning hyperparameters. In addition, the proposed
framework has several more general conceptual implications for outlier
detection which will be summarized in the following.

\hypertarget{sec:disc:implications}{%
\subsection{Implications}\label{sec:disc:implications}}

\textbf{Outlier taxonomy} We propose a clear taxonomy to distinguish
between frequently interchangeably used terms \emph{anomalies} and
\emph{outliers} in a canonical way: we regard \emph{anomalies} as
observations from a different data generating process than the majority
of the data (i.e.~as observations that are on \(\Man\) but not on
\(\Min\)), which can be more precisely identified as \emph{structural}
outliers. Recall that Zimek and Filzmoser
(\protect\hyperlink{ref-zimek2018there}{2018, p. 10}) refer to such
observations as ``real'' outliers that need to be distinguished from
``observations which are in the extremes of the model distribution''. On
the other hand, regarding \emph{outliers} as observations from
low-density regions of the underlying ``normal'' data manifold \(\Min\),
they can be more precisely identified as \emph{distributional} outliers.
Based on our reading of the literature, this distinction is usually not
made explicit. Since there is rarely a practical reason to assume that a
given data set contains only \textit{distributional} or only
\textit{structural} outliers, some of the confusion surrounding the
topic (\protect\hyperlink{ref-goldstein2016comparative}{Goldstein \&
Uchida, 2016}; \protect\hyperlink{ref-unwin2019multivariate}{Unwin,
2019}; \protect\hyperlink{ref-zimek2018there}{Zimek \& Filzmoser, 2018})
might be because such conceptual differences have not been made
sufficiently clear. As outlined, the concept of structural difference is
very general. For example, structural differences in functional data may
appear as shape anomalies in data mainly characterized by vertical shift
variation (see Fig. \ref{fig:outtypes} A) or as vertical shift anomalies
in data dominated by shape variation, as phase anomalies in data with
magnitude variation or magnitude anomalies in data with phase variation,
etc.\\
In real unlabeled data, there may not always be a clear distinction
between somewhat structurally anomalous observations with
``off-manifold'' embeddings and merely distributionally outlying
observations with embeddings on the periphery of the data manifold, as
in the ECG data in Figure \ref{fig:fda-image-real} A. Nevertheless, the
theoretical distinction between these two kinds of outliers adds
conceptual clarity even if the practical application of the categories
may not be straightforward.\\
\textbf{Curse of dimensionality} As outlined in section
\ref{sec:prelims:scope}, outlier detection is often reported to suffer
from the curse of dimensionality. For example, Goldstein \& Uchida
(\protect\hyperlink{ref-goldstein2016comparative}{2016}) show that most
outlier detection methods under consideration break down or perform
poorly in a data set with 400 dimensions and conclude that unsupervised
outlier detection is not possible in such high dimensions. Some
{[}Aggarwal (\protect\hyperlink{ref-aggarwal2017outlier}{2017}); e.g.{]}
attribute this to the fundamental problem that distance functions can
lose their discriminating power in high dimensions
(\protect\hyperlink{ref-beyer1999nearest}{Beyer et al., 1999}), which is
linked to the concentration of measure effect
(\protect\hyperlink{ref-pestov2000geometry}{Pestov, 2000}). However,
this effect occurs only under fairly specific conditions
(\protect\hyperlink{ref-zimek2012survey}{Zimek et al., 2012}), which
means that outlier detection does not have to be affected by the curse
of dimensionality: In addition to the effects of dependency structures
and signal-to-noise ratios
(\protect\hyperlink{ref-zimek2012survey}{Zimek et al., 2012}), the
necessary conditions for concentration of measure are not fulfilled if
the intrinsic dimensionality of the data is smaller than the actually
observed dimensionality, or if the data is distributed in clusters that
are relatively well separable
(\protect\hyperlink{ref-beyer1999nearest}{Beyer et al., 1999}). Exactly
these two characteristics are reflected in our framework in the form of
(1) the manifold assumption, which implies low-ish intrinsic
dimensionality, and (2) the assumption that structural outliers come
from different manifolds than the rest of the data, i.e., from different
``clusters'' in \(\hdspace\). This has two important consequences: First
of all, the geometric perspective our framework is based on makes these
important aspects for outlier detection in high-dimensional data
explicit, while a purely probabilistic perspective obscures them.
Secondly, it mitigates many of the problems associated with
high-dimensional outlier detection: any outlier detection method that
performs well in low dimensions becomes -- in principle -- applicable in
nominally high-dimensional and/or complex non-tabular data when applied
to suitable low-dimensional embedding coordinates. In addition, our
results show that outlier sub-structure, specifically the differences
between distributional and structural outliers, can be detected and
visualized with manifold methods. This opens new possibilities for
descriptive and exploratory analyses:\\
\textbf{Visualizability of outlier characteristics} If the embeddings
provided by manifold methods are restricted to two or three dimensions,
they also provide easily accessible visualizations of the data. In fact,
manifold learning is often used in applications specifically to find
two- or three-dimensional visualizations reflecting the essential
intrinsic structure of the high-dimensional data as faithfully as
possible. Consequently, structural and distributional outliers, which
are rather glaring data characteristics if the manifolds are well
separable, can often be separated clearly even in two- or
three-dimensional representations as long as the embedding is
(approximately) isometric with respect to a suitable dissimilarity
measure. This is specifically important for complex non-tabular or
high-dimensional data types such as images or graphs, where at most a
few observations can be visualized and perceived simultaneously. In the
same vein, substructures and notions of data depth are reflected in the
embeddings, making the approach also useful as an exploration tool for
settings with unclear structure.\\
\textbf{Generalizability} Since the central building block of the
proposed framework is to capture the metric structure of data sets using
distance measures, the framework is very general and applicable to any
data type for which distance metrics are available. In Section
\ref{sec:exps:qual}, we illustrated this generalizability using
high-dimensional as well as non-tabular data; in particular, we applied
it to functional, curve, graph, and image data. This also makes the
framework very flexible as one can make use of non-standard and
customized dissimilarity measures to emphasize the relevant structural
differences in specific situations based on domain knowledge:
Representing image data as vectors of pixel intensities, we computed
distances between those vectors, for example. Dissimilarities between
different graphs were captured, for example, by constructing their graph
Laplacians and computing Frobenius distances between them, and we used a
specific elastic depth distance for the spiral curve data as suggested
by earlier results in Steyer et al.
(\protect\hyperlink{ref-steyer2021elastic}{2021}).

\hypertarget{sec:discussion:limitations}{%
\subsection{Limitations and outlook}\label{sec:discussion:limitations}}

\color{black}

If in an exploratory setting, observations appear clearly separated in
the (first few) embedding dimensions, we can be sure they are structural
outliers. Note that if \(D\)-dimensional data actually live in a
\(d'\)-dimensional subspace, constructing a \(d'\)-dimensional embedding
with MDS based on \(L_2\) distances will lead to an embedding with a
distance matrix exactly matching the distance matrix in the
\(D\)-dimensional space, i.e.~MDS is isometric by design. If other than
\(L_2\) distances are used this still holds approximately
(\protect\hyperlink{ref-youngDiscussionSetPoints1938}{Young \&
Householder, 1938}; see also
\protect\hyperlink{ref-cox2008multidimensional}{Cox \& Cox, 2008};
\protect\hyperlink{ref-torgersonMultidimensionalScalingTheory1952}{Torgerson,
1952}). \color{black} Note that this is an important difference from many
other dimension reduction methods. For example, UMAP is based on a local
connectivity constraint
(\protect\hyperlink{ref-mcinnesUMAPUniformManifold2020}{McInnes et al.,
2020}) which ensures that each point is at least connected to its
nearest neighbor and which runs counter to a reliable embedding of
structural outliers. In addition, more sophisticated methods require
parameter tuning for any given setting, which is inherently difficult
for unsupervised tasks, and it is not always clear how to tune other
embedding methods so that they yield (approximately) isometric
embeddings.\\
\color{black} Clear structural outliers are the source of large
variation in data sets with low intrinsic dimensionality. Since MDS
embedding dimensions are sorted according to the decreasing variation,
they will be reflected in the first few embedding dimensions. It may be
that some of the distributional outliers are masked due to projection if
the embedding dimension \(d\) is smaller than \(d'\) but following Zimek
\& Filzmoser (\protect\hyperlink{ref-zimek2018there}{2018}) we consider
faithfully reflecting structural outliers more important. However,
inliers, i.e.~observations on \(\Min\), may show large ``within class''
variation and/or may be spread over several disconnected clusters in
some situations. For example, object images on \(\Min\), which are
structurally similar in terms of the depicted objects' shape, may vary
in rotation, scale, or location, and may have different colors or
textures. In functional data, observations on \(\Min\) may show phase
and amplitude variation and form clusters due to different shapes. In
such settings, \(\Min\) can yield complex substructure and highly
dispersed observations and it may be hard to distinguish whether
separable structures observed in embeddings are due to groups of
homogeneous structural outliers or due to multimodality in \(\Min\) in
which some modes are sparsely sampled. Moreover, in such cases, the
dispersion of \(\Min\) accounts for large parts of the data's
variability, and two- or three-dimensional MDS embeddings may not be
sufficient to also faithfully represent structural outliers, since MDS
embedding vectors are sorted decreasingly by explained ``variance''.
However, this does not mean that structural outliers are not necessarily
separable. Instead, they appear as outliers in higher embedding
dimensions, requiring higher order embeddings to reflect the outlier
structure. \color{black} That means, if in an exploratory setting, there
are no clearly separated observations in the (first few) embedding
dimensions, there are either no clear structural outliers or they appear
in later embedding dimensions if there are sources in \(M_c\) that
induce more variation than the structural outlier. For example, objects
in images may be structurally different in texture but not in color,
orientation, and scale. In such a case -- all observations differ in
color, orientation, and scale but only some observations in texture --,
these other aspects can induce large variation within observations on
\(M_c\), and the structural difference in texture is loaded on latter
embedding dimensions. In such a situation, one can use scatterplot
matrices and \emph{Scagnostics} (scatterplot diagnostics,
\protect\hyperlink{ref-wilkinsonGraphtheoreticScagnostics2005}{Wilkinson
et al., 2005}) for visual inspection. In addition, one can check out the
kurtosis of the LOF scores in different embedding dimensions or
\emph{high contrast subspaces for density-based outlier ranking} (HiCS,
\protect\hyperlink{ref-kellerHiCSHighContrast2012}{Keller et al.,
2012}), to find pairs of dimensions that are ``interesting'' in terms of
structural outliers. Moreover, \color{black} techniques from multi-view
learning such as ``distance-learning from multiple views'' may likely
yield better results, because different structures (e.g.~structure
induced by color vs structure induced by texture) should be ``treated
separately as they are \emph{semantically} different''
(\protect\hyperlink{ref-zimek2015blind}{Zimek \& Vreeken, 2015, p.
128}). Note, however, that suitable inductive biases can also be brought
to bear in our framework fairly easily. If substantial considerations
suggest that specific structural aspects are important, specifying
dissimilarity metrics focused on these aspects allows to emphasize the
relevant differences. For example, if isolated outliers in functional
data (i.e.~functions which yield outlying behavior only over small parts
of the domain such as isolated peaks) are of most interest, higher order
\(L_p\) metrics such as \(L_{10}\) will be much more sensitive to such
structural differences than general \(L_2\) distances. If phase
variation should be ignored, the unnormalized \(L_1\)-Wasserstein or the
Dynamic Time Warping (DTW) distance can be used. Such problem-specific
distance measures can reduce the number of MDS embedding dimensions
necessary for faithful embeddings of structural outliers
(\protect\hyperlink{ref-herrmann2021geometric}{Herrmann \& Scheipl,
2021}). In future work, we will investigate these aspects and possible
extensions w.r.t. to multi-view learning approaches. Moreover, we will
elaborate more on the specifics of other data types, in particular,
image data.

\hypertarget{sec:conclusion}{%
\section{Conclusion}\label{sec:conclusion}}

In conclusion, our illustration suggests that the proposed geometric
conceptualization, which distinguishes \emph{distributional} and
\emph{structural} outliers on a general level, provides a more precise
terminology and shows that outlier detection in high-dimensional and
complex non-tabular data does need to be specifically challenging per
se. Convincing results could be achieved in a wide range of settings and
data types by a combination of the simple methods MDS for dimension
reduction and visualization and LOF for outlier scoring. We hope that
the proposed framework contributes to a better understanding of
unsupervised outlier detection and provides some guidance to
practitioners as well as methodological researchers in this regard.

\section*{Funding}

This work has been funded by the German Federal Ministry of Education
and Research (BMBF) under Grant No.~01IS18036A. The authors of this work
take full responsibility for its content.

\section*{Acknowledgment}

The authors thank Almond Stöcker for his helpful advice regarding the
spiral curve data.

\section*{Conflict of interest}

The authors have declared no conflicts of interest for this article.

\section*{Data availability statement}

The data and code to reproduce the findings of this study are openly
available on GitHub at:
\url{https://github.com/HerrMo/geo-outlier-framework}

\section*{References}

\hypertarget{refs}{}
\begin{CSLReferences}{1}{0}
\leavevmode\vadjust pre{\hypertarget{ref-aggarwal2017outlier}{}}%
Aggarwal, C. C. (2017). \emph{Outlier analysis} (2nd ed.). Springer.
\url{https://doi.org/10.1007/978-3-319-47578-3}

\leavevmode\vadjust pre{\hypertarget{ref-aggarwal2001outlier}{}}%
Aggarwal, C. C., \& Yu, P. S. (2001). Outlier detection for high
dimensional data. \emph{SIGMOD Rec.}, \emph{30}(2).
\url{https://doi.org/10.1145/376284.375668}

\leavevmode\vadjust pre{\hypertarget{ref-ali2019timecluster}{}}%
Ali, M., Jones, M. W., Xie, X., \& Williams, M. (2019). Time{C}luster:
Dimension reduction applied to temporal data for visual analytics.
\emph{The Visual Computer}, \emph{35}(6), 1013--1026.
\url{https://doi.org/10.1007/s00371-019-01673-y}

\leavevmode\vadjust pre{\hypertarget{ref-alty2017use}{}}%
Alty, J., Cosgrove, J., Thorpe, D., \& Kempster, P. (2017). How to use
pen and paper tasks to aid tremor diagnosis in the clinic.
\emph{Practical Neurology}, \emph{17}(6), 456--463.
\url{https://doi.org/10.1136/practneurol-2017-001719}

\leavevmode\vadjust pre{\hypertarget{ref-anderson1935irises}{}}%
Anderson, E. (1935). The irises of the gaspé peninsula. \emph{Bull Am
Iris Soc}, \emph{59}, 2--5.

\leavevmode\vadjust pre{\hypertarget{ref-azcorra2018unsupervised}{}}%
Azcorra, A., Chiroque, L. F., Cuevas, R., Anta, A. F., Laniado, H.,
Lillo, R. E., Romo, J., \& Sguera, C. (2018). Unsupervised scalable
statistical method for identifying influential users in online social
networks. \emph{Scientific Reports}, \emph{8}(1), 1--7.
\url{https://doi.org/10.1038/s41598-018-24874-2}

\leavevmode\vadjust pre{\hypertarget{ref-beckmanOutlier1983}{}}%
Beckman, R. J., \& Cook, R. D. (1983). Outlier \ldots{} \ldots{} \ldots.
s. \emph{Technometrics}, \emph{25}(2), 119--149.
\url{https://doi.org/10.1080/00401706.1983.10487840}

\leavevmode\vadjust pre{\hypertarget{ref-belkin2003laplacian}{}}%
Belkin, M., \& Niyogi, P. (2003). Laplacian eigenmaps for dimensionality
reduction and data representation. \emph{Neural Computation},
\emph{15}(6), 1373--1396.
\url{https://doi.org/10.1162/089976603321780317}

\leavevmode\vadjust pre{\hypertarget{ref-beyer1999nearest}{}}%
Beyer, K., Goldstein, J., Ramakrishnan, R., \& Shaft, U. (1999). When is
{``nearest neighbor''} meaningful? In C. Beeri \& P. Buneman (Eds.),
\emph{Database theory --- ICDT'99} (pp. 217--235). Springer Berlin
Heidelberg. \url{https://doi.org/10.1007/3-540-49257-7_15}

\leavevmode\vadjust pre{\hypertarget{ref-breunig2000lof}{}}%
Breunig, M. M., Kriegel, H.-P., Ng, R. T., \& Sander, J. (2000). {LOF}:
Identifying density-based local outliers. \emph{SIGMOD Rec.},
\emph{29}(2), 93--104. \url{https://doi.org/10.1145/335191.335388}

\leavevmode\vadjust pre{\hypertarget{ref-campos2016evaluation}{}}%
Campos, G. O., Zimek, A., Sander, J., Campello, R. J., Micenková, B.,
Schubert, E., Assent, I., \& Houle, M. E. (2016). On the evaluation of
unsupervised outlier detection: Measures, datasets, and an empirical
study. \emph{Data Mining and Knowledge Discovery}, \emph{30}(4),
891--927. \url{https://doi.org/10.1007/s10618-015-0444-8}

\leavevmode\vadjust pre{\hypertarget{ref-clemenccon2013scoring}{}}%
Clémençon, S., \& Jakubowicz, J. (2013). Scoring anomalies: A
{M}-estimation formulation. In C. M. Carvalho \& P. Ravikumar (Eds.),
\emph{Proceedings of the sixteenth international conference on
artificial intelligence and statistics} (Vol. 31, pp. 659--667). PMLR.
\url{https://proceedings.mlr.press/v31/clemencon13a.html}

\leavevmode\vadjust pre{\hypertarget{ref-coifman2006diffusion}{}}%
Coifman, R. R., \& Lafon, S. (2006). Diffusion maps. \emph{Applied and
Computational Harmonic Analysis}, \emph{21}(1), 5--30.
\url{https://doi.org/10.1016/j.acha.2006.04.006}

\leavevmode\vadjust pre{\hypertarget{ref-cox2008multidimensional}{}}%
Cox, M. A. A., \& Cox, T. F. (2008). Multidimensional {Scaling}. In C.
Chen, W. Härdle, \& A. Unwin (Eds.), \emph{Handbook of {Data
Visualization}} (pp. 315--347). {Springer}.
\url{https://doi.org/10.1007/978-3-540-33037-0_14}

\leavevmode\vadjust pre{\hypertarget{ref-dai2020functional}{}}%
Dai, W., Mrkvička, T., Sun, Y., \& Genton, M. G. (2020). Functional
outlier detection and taxonomy by sequential transformations.
\emph{Computational Statistics \& Data Analysis}, \emph{149}, 106960.
\url{https://doi.org/10.1016/j.csda.2020.106960}

\leavevmode\vadjust pre{\hypertarget{ref-dau2019ucr}{}}%
Dau, H. A., Bagnall, A., Kamgar, K., Yeh, C.-C. M., Zhu, Y., Gharghabi,
S., Ratanamahatana, C. A., \& Keogh, E. (2019). The {UCR} time series
archive. \emph{IEEE/CAA Journal of Automatica Sinica}, \emph{6}(6),
1293--1305. \url{https://doi.org/10.1109/JAS.2019.1911747}

\leavevmode\vadjust pre{\hypertarget{ref-Dua2019}{}}%
Dua, D., \& Graff, C. (2017). \emph{{UCI} machine learning repository}.
University of California, Irvine, School of Information; Computer
Sciences. \url{http://archive.ics.uci.edu/ml}

\leavevmode\vadjust pre{\hypertarget{ref-febrero2012fdausc}{}}%
Febrero-Bande, M., \& Oviedo de la Fuente, M. (2012). Statistical
computing in functional data analysis: The {R} package {fda.usc}.
\emph{Journal of Statistical Software}, \emph{51}(4), 1--28.
\url{https://doi.org/10.18637/jss.v051.i04}

\leavevmode\vadjust pre{\hypertarget{ref-fisher1936use}{}}%
Fisher, R. A. (1936). The use of multiple measurements in taxonomic
problems. \emph{Ann Eugen}, \emph{7}(2), 179--188.

\leavevmode\vadjust pre{\hypertarget{ref-fritsch2012detecting}{}}%
Fritsch, V., Varoquaux, G., Thyreau, B., Poline, J.-B., \& Thirion, B.
(2012). Detecting outliers in high-dimensional neuroimaging datasets
with robust covariance estimators. \emph{Medical Image Analysis},
\emph{16}(7), 1359--1370.
\url{https://doi.org/10.1016/j.media.2012.05.002}

\leavevmode\vadjust pre{\hypertarget{ref-goldberger2000physiobank}{}}%
Goldberger, A. L., Amaral, L. A., Glass, L., Hausdorff, J. M., Ivanov,
P. C., Mark, R. G., Mietus, J. E., Moody, G. B., Peng, C.-K., \&
Stanley, H. E. (2000). Physio{B}ank, {P}hysio{T}oolkit, and
{P}hysio{N}et: Components of a new research resource for complex
physiologic signals. \emph{Circulation}, \emph{101}(23), e215--e220.
\url{https://doi.org/10.1161/01.CIR.101.23.e215}

\leavevmode\vadjust pre{\hypertarget{ref-goldstein2016comparative}{}}%
Goldstein, M., \& Uchida, S. (2016). A comparative evaluation of
unsupervised anomaly detection algorithms for multivariate data.
\emph{PloS One}, \emph{11}(4), e0152173.
\url{https://doi.org/10.1371/journal.pone.0152173}

\leavevmode\vadjust pre{\hypertarget{ref-guanNovelIntrinsicMeasure2021}{}}%
Guan, S., \& Loew, M. (2021). A {Novel Intrinsic Measure} of {Data
Separability}. \emph{arXiv:2109.05180 {[}Cs, Math, Stat{]}}.
\url{http://arxiv.org/abs/2109.05180}

\leavevmode\vadjust pre{\hypertarget{ref-hernandez2106kernel}{}}%
Hernández, N., \& Muñoz, A. (2016). Kernel {D}epth {M}easures for
{F}unctional {D}ata with {A}pplication to {O}utlier {D}etection. In A.
E. P. Villa, P. Masulli, \& A. J. Pons Rivero (Eds.), \emph{Artificial
neural networks and machine learning -- {ICANN} 2016} (pp. 235--242).
Springer, Cham. \url{https://doi.org/10.1007/978-3-319-44781-0_28}

\leavevmode\vadjust pre{\hypertarget{ref-herrmann2021geometric}{}}%
Herrmann, M., \& Scheipl, F. (2021). A geometric perspective on
functional outlier detection. \emph{Stats}, \emph{4}(4), 971--1011.
\url{https://doi.org/10.3390/stats4040057}

\leavevmode\vadjust pre{\hypertarget{ref-isenkul2014improved}{}}%
Isenkul, M., Sakar, B., Kursun, O., et al. (2014). Improved spiral test
using digitized graphics tablet for monitoring parkinson's disease.
\emph{The 2nd International Conference on e-Health and Telemedicine
(ICEHTM-2014)}, \emph{5}, 171--175.

\leavevmode\vadjust pre{\hypertarget{ref-kamalov2020outlier}{}}%
Kamalov, F., \& Leung, H. H. (2020). Outlier detection in high
dimensional data. \emph{Journal of Information \& Knowledge Management},
\emph{19}(01), 2040013. \url{https://doi.org/10.1142/S0219649220400134}

\leavevmode\vadjust pre{\hypertarget{ref-kandanaarachchi2020dimension}{}}%
Kandanaarachchi, S., \& Hyndman, R. J. (2020). Dimension reduction for
outlier detection using {DOBIN}. \emph{Journal of Computational and
Graphical Statistics}, 1--16.
\url{https://doi.org/10.1080/10618600.2020.1807353}

\leavevmode\vadjust pre{\hypertarget{ref-kellerHiCSHighContrast2012}{}}%
Keller, F., Muller, E., \& Bohm, K. (2012). {HiCS}: {High Contrast
Subspaces} for {Density-Based Outlier Ranking}. \emph{2012 {IEEE} 28th
{International Conference} on {Data Engineering}}, 1037--1048.
\url{https://doi.org/10.1109/ICDE.2012.88}

\leavevmode\vadjust pre{\hypertarget{ref-lee2007nonlinear}{}}%
Lee, J. A., \& Verleysen, M. (2007). \emph{Nonlinear {Dimensionality}
{Reduction}} (1st ed.). Springer Science \& Business Media.
\url{https://doi.org/10.1007/978-0-387-39351-3}

\leavevmode\vadjust pre{\hypertarget{ref-loperfido2020kurtosis}{}}%
Loperfido, N. (2020). Kurtosis-based projection pursuit for outlier
detection in financial time series. \emph{The European Journal of
Finance}, \emph{26}(2-3), 142--164.
\url{https://doi.org/10.1080/1351847X.2019.1647864}

\leavevmode\vadjust pre{\hypertarget{ref-ma2011manifold}{}}%
Ma, Y., \& Fu, Y. (Eds.). (2011). \emph{Manifold learning theory and
applications} (1st ed.). CRC press.
\url{https://doi.org/doi.org/10.1201/b11431}

\leavevmode\vadjust pre{\hypertarget{ref-maaten2008visualizing}{}}%
Maaten, L. van der, \& Hinton, G. (2008). Visualizing data using
t-{SNE}. \emph{Journal of Machine Learning Research}, \emph{9}(86),
2579--2605. \url{http://jmlr.org/papers/v9/vandermaaten08a.html}

\leavevmode\vadjust pre{\hypertarget{ref-marques2020internal}{}}%
Marques, H. O., Campello, R. J., Sander, J., \& Zimek, A. (2020).
Internal evaluation of unsupervised outlier detection. \emph{ACM
Transactions on Knowledge Discovery from Data (TKDD)}, \emph{14}(4),
1--42. \url{https://doi.org/10.1145/3394053}

\leavevmode\vadjust pre{\hypertarget{ref-mcinnesUMAPUniformManifold2020}{}}%
McInnes, L., Healy, J., \& Melville, J. (2020). {UMAP}: {Uniform
Manifold Approximation} and {Projection} for {Dimension Reduction}.
\emph{arXiv:1802.03426 {[}Cs, Stat{]}}.
\url{http://arxiv.org/abs/1802.03426}

\leavevmode\vadjust pre{\hypertarget{ref-mcinnes2018umap}{}}%
McInnes, L., Healy, J., \& Melville, J. (2018). \emph{{UMAP}: Uniform
manifold approximation and projection for dimension reduction}. arXiv.
\url{https://doi.org/10.48550/ARXIV.1802.03426}

\leavevmode\vadjust pre{\hypertarget{ref-mordohaiDimensionalityEstimationManifold2010}{}}%
Mordohai, P., \& Medioni, G. (2010). Dimensionality {Estimation},
{Manifold Learning} and {Function Approximation} using {Tensor Voting}.
\emph{Journal of Machine Learning Research}, \emph{11}(12), 411--450.
\url{http://jmlr.org/papers/v11/mordohai10a.html}

\leavevmode\vadjust pre{\hypertarget{ref-munoz2004one}{}}%
Muñoz, A., \& Moguerza, J. M. (2004). One-class support vector machines
and density estimation: The precise relation. In A. Sanfeliu, J. F.
Martínez Trinidad, \& J. A. Carrasco Ochoa (Eds.), \emph{Progress in
pattern recognition, image analysis and applications} (pp. 216--223).
Springer Berlin Heidelberg.
\url{https://doi.org/10.1007/978-3-540-30463-0_27}

\leavevmode\vadjust pre{\hypertarget{ref-coil20}{}}%
Nane, S., Nayar, S., \& Murase, H. (1996). Columbia object image
library: COIL-20. \emph{Dept. Comp. Sci., Columbia University, New York,
Tech. Rep}.

\leavevmode\vadjust pre{\hypertarget{ref-navarro2021high}{}}%
Navarro-Esteban, P., \& Cuesta-Albertos, J. A. (2021). High-dimensional
outlier detection using random projections. \emph{TEST}, \emph{30}(4),
908--934. \url{https://doi.org/10.1007/s11749-020-00750-y}

\leavevmode\vadjust pre{\hypertarget{ref-niyogiTopologicalViewUnsupervised2011}{}}%
Niyogi, P., Smale, S., \& Weinberger, S. (2011). A {Topological View} of
{Unsupervised Learning} from {Noisy Data}. \emph{SIAM J. Comput.},
\emph{40}, 646--663. \url{https://doi.org/10.1137/090762932}

\leavevmode\vadjust pre{\hypertarget{ref-pang2018learning}{}}%
Pang, G., Cao, L., Chen, L., \& Liu, H. (2018). Learning representations
of ultrahigh-dimensional data for random distance-based outlier
detection. \emph{Proceedings of the 24th ACM SIGKDD International
Conference on Knowledge Discovery \& Data Mining}, 2041--2050.
\url{https://doi.org/10.1145/3219819.3220042}

\leavevmode\vadjust pre{\hypertarget{ref-pestov2000geometry}{}}%
Pestov, V. (2000). On the geometry of similarity search: Dimensionality
curse and concentration of measure. \emph{Information Processing
Letters}, \emph{73}(1-2), 47--51.
\url{https://doi.org/10.1016/S0020-0190(99)00156-8}

\leavevmode\vadjust pre{\hypertarget{ref-polonik1997minimum}{}}%
Polonik, W. (1997). Minimum volume sets and generalized quantile
processes. \emph{Stochastic Processes and Their Applications},
\emph{69}(1), 1--24. \url{https://doi.org/10.1016/S0304-4149(97)00028-8}

\leavevmode\vadjust pre{\hypertarget{ref-ramsay2005functional}{}}%
Ramsay, J. O., \& Silverman, B. W. (2005). \emph{Functional data
analysis} (2nd ed). Springer. \url{https://doi.org/10.1007/b98888}

\leavevmode\vadjust pre{\hypertarget{ref-rebbapragada2009finding}{}}%
Rebbapragada, U., Protopapas, P., Brodley, C. E., \& Alcock, C. (2009).
Finding anomalous periodic time series. \emph{Machine Learning},
\emph{74}(3), 281--313. \url{https://doi.org/10.1007/s10994-008-5093-3}

\leavevmode\vadjust pre{\hypertarget{ref-ren2017projection}{}}%
Ren, H., Chen, N., \& Zou, C. (2017). Projection-based outlier detection
in functional data. \emph{Biometrika}, \emph{104}(2), 411--423.
\url{https://doi.org/10.1093/biomet/asx012}

\leavevmode\vadjust pre{\hypertarget{ref-ro2015outlier}{}}%
Ro, K., Zou, C., Wang, Z., \& Yin, G. (2015). Outlier detection for
high-dimensional data. \emph{Biometrika}, \emph{102}(3), 589--599.
\url{https://doi.org/10.1093/biomet/asv021}

\leavevmode\vadjust pre{\hypertarget{ref-rousseeuw2005robust}{}}%
Rousseeuw, P. J., \& Leroy, A. M. (2005). \emph{Robust regression and
outlier detection}. John Wiley \& Sons.
\url{https://doi.org/10.1002/0471725382}

\leavevmode\vadjust pre{\hypertarget{ref-roweis2000nonlinear}{}}%
Roweis, S. T., \& Saul, L. K. (2000). Nonlinear dimensionality reduction
by locally linear embedding. \emph{Science}, \emph{290}(5500),
2323--2326. \url{https://doi.org/10.1126/science.290.5500.2323}

\leavevmode\vadjust pre{\hypertarget{ref-scott2006learning}{}}%
Scott, C. D., \& Nowak, R. D. (2006). Learning minimum volume sets.
\emph{The Journal of Machine Learning Research}, \emph{7}(24), 665--704.
\url{http://jmlr.org/papers/v7/scott06a.html}

\leavevmode\vadjust pre{\hypertarget{ref-souvenirManifoldClustering2005}{}}%
Souvenir, R., \& Pless, R. (2005). Manifold clustering. \emph{Tenth
{IEEE International Conference} on {Computer Vision} ({ICCV}'05)
{Volume} 1}, 648--653 Vol. 1.
\url{https://doi.org/10.1109/ICCV.2005.149}

\leavevmode\vadjust pre{\hypertarget{ref-steyer2021elastic}{}}%
Steyer, L., Stöcker, A., \& Greven, S. (2021). \emph{Elastic analysis of
irregularly or sparsely sampled curves}. arXiv.
\url{https://doi.org/10.48550/ARXIV.2104.11039}

\leavevmode\vadjust pre{\hypertarget{ref-streetNuclearFeatureExtraction1993}{}}%
Street, W. N., Wolberg, W. H., \& Mangasarian, O. L. (1993). Nuclear
feature extraction for breast tumor diagnosis. \emph{Biomedical {Image
Processing} and {Biomedical Visualization}}, \emph{1905}, 861--870.
\url{https://doi.org/10.1117/12.148698}

\leavevmode\vadjust pre{\hypertarget{ref-tenenbaum2000global}{}}%
Tenenbaum, J. B., Silva, V. de, \& Langford, J. C. (2000). A {Global}
{Geometric} {Framework} for {Nonlinear} {Dimensionality} {Reduction}.
\emph{Science}, \emph{290}(5500), 2319--2323.
\url{https://doi.org/10.1126/science.290.5500.2319}

\leavevmode\vadjust pre{\hypertarget{ref-thudumu2020comprehensive}{}}%
Thudumu, S., Branch, P., Jin, J., \& Singh, J. J. (2020). A
comprehensive survey of anomaly detection techniques for high
dimensional big data. \emph{Journal of Big Data}, \emph{7}(1), 1--30.
\url{https://doi.org/10.1186/s40537-020-00320-x}

\leavevmode\vadjust pre{\hypertarget{ref-toivola2010novelty}{}}%
Toivola, J., Prada, M. A., \& Hollmén, J. (2010). Novelty detection in
projected spaces for structural health monitoring. In P. R. Cohen, N. M.
Adams, \& M. R. Berthold (Eds.), \emph{Advances in intelligent data
analysis IX} (pp. 208--219). Springer Berlin Heidelberg.
\url{https://doi.org/10.1007/978-3-642-13062-5_20}

\leavevmode\vadjust pre{\hypertarget{ref-torgersonMultidimensionalScalingTheory1952}{}}%
Torgerson, W. S. (1952). Multidimensional scaling: {I}. {Theory} and
method. \emph{Psychometrika}, \emph{17}(4), 401--419.
\url{https://doi.org/10.1007/BF02288916}

\leavevmode\vadjust pre{\hypertarget{ref-unwin2019multivariate}{}}%
Unwin, A. (2019). Multivariate outliers and the {O}3 plot. \emph{Journal
of Computational and Graphical Statistics}, \emph{28}(3), 635--643.
\url{https://doi.org/10.1080/10618600.2019.1575226}

\leavevmode\vadjust pre{\hypertarget{ref-wilkinsonGraphtheoreticScagnostics2005}{}}%
Wilkinson, L., Anand, A., \& Grossman, R. (2005). Graph-theoretic
scagnostics. \emph{{IEEE Symposium} on {Information Visualization},
2005. {INFOVIS} 2005.}, 157--164.
\url{https://doi.org/10.1109/INFVIS.2005.1532142}

\leavevmode\vadjust pre{\hypertarget{ref-xie2017geometric}{}}%
Xie, W., Kurtek, S., Bharath, K., \& Sun, Y. (2017). A {G}eometric
{A}pproach to {V}isualization of {V}ariability in {F}unctional data.
\emph{Journal of the American Statistical Association}, \emph{112}(519),
979--993. \url{https://doi.org/10.1080/01621459.2016.1256813}

\leavevmode\vadjust pre{\hypertarget{ref-xu2018comparison}{}}%
Xu, X., Liu, H., Li, L., \& Yao, M. (2018). A comparison of outlier
detection techniques for high-dimensional data. \emph{International
Journal of Computational Intelligence Systems}, \emph{11}(1), 652--662.
\url{https://doi.org/10.2991/ijcis.11.1.50}

\leavevmode\vadjust pre{\hypertarget{ref-youngDiscussionSetPoints1938}{}}%
Young, G., \& Householder, A. S. (1938). Discussion of a set of points
in terms of their mutual distances. \emph{Psychometrika}, \emph{3}(1),
19--22. \url{https://doi.org/10.1007/BF02287916}

\leavevmode\vadjust pre{\hypertarget{ref-zhang2006anomaly}{}}%
Zhang, J., \& Zulkernine, M. (2006). Anomaly based network intrusion
detection with unsupervised outlier detection. \emph{2006 IEEE
International Conference on Communications}, \emph{5}, 2388--2393.
\url{https://doi.org/10.1109/ICC.2006.255127}

\leavevmode\vadjust pre{\hypertarget{ref-zimek2018there}{}}%
Zimek, A., \& Filzmoser, P. (2018). There and back again: Outlier
detection between statistical reasoning and data mining algorithms.
\emph{Wiley Interdisciplinary Reviews: Data Mining and Knowledge
Discovery}, \emph{8}(6), e1280. \url{https://doi.org/10.1002/widm.1280}

\leavevmode\vadjust pre{\hypertarget{ref-zimek2012survey}{}}%
Zimek, A., Schubert, E., \& Kriegel, H.-P. (2012). A survey on
unsupervised outlier detection in high-dimensional numerical data.
\emph{Statistical Analysis and Data Mining: The ASA Data Science
Journal}, \emph{5}(5), 363--387. \url{https://doi.org/10.1002/sam.11161}

\leavevmode\vadjust pre{\hypertarget{ref-zimek2015blind}{}}%
Zimek, A., \& Vreeken, J. (2015). The blind men and the elephant: On
meeting the problem of multiple truths in data from clustering and
pattern mining perspectives. \emph{Machine Learning}, \emph{98}(1),
121--155. \url{https://doi.org/10.1007/s10994-013-5334-y}

\end{CSLReferences}

\newpage

\appendix

\hypertarget{sec:app-1}{%
\section{Example visualizations of the data used in the quantitative
experiments}\label{sec:app-1}}

\begin{figure}
\centering
\includegraphics{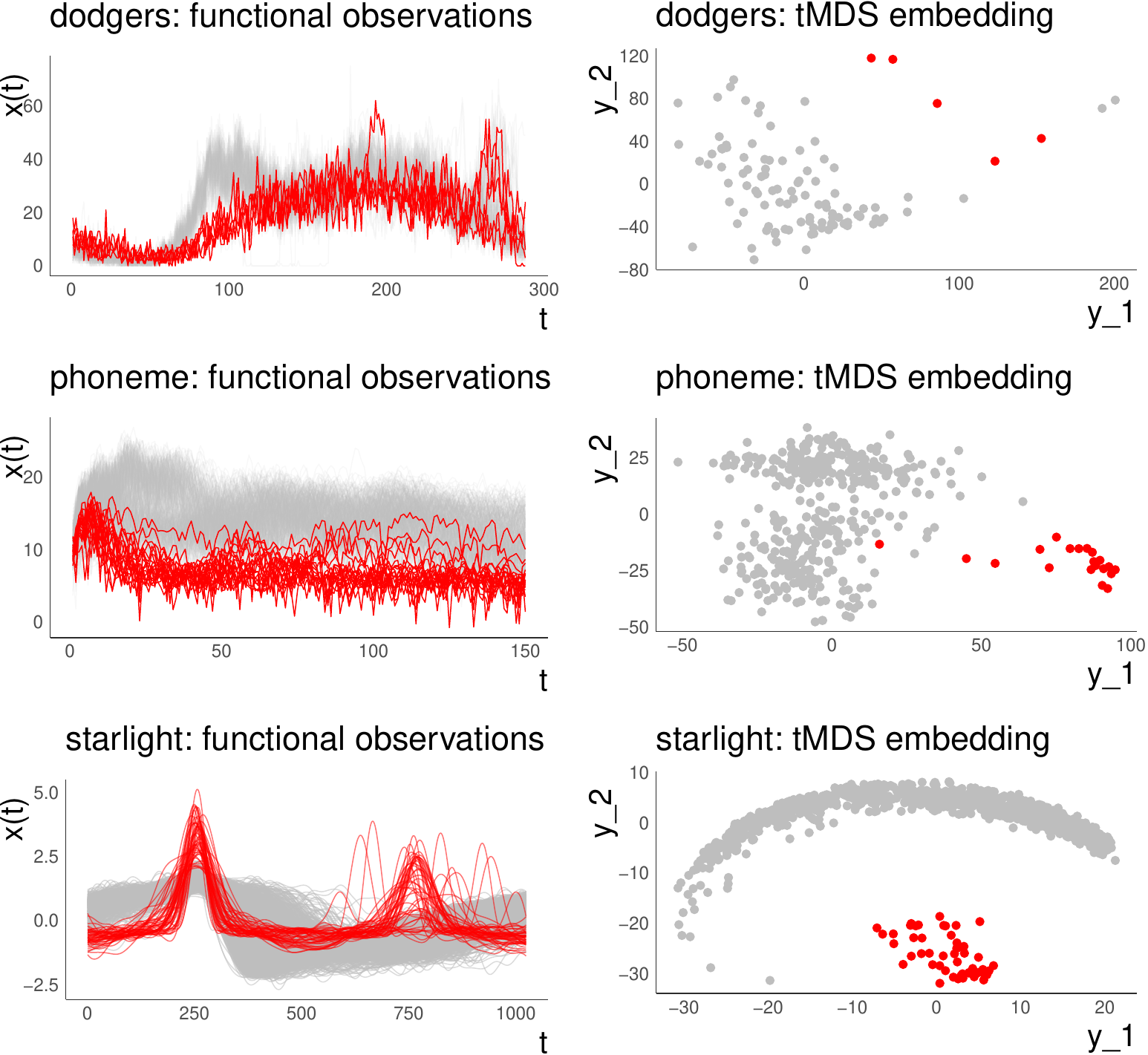}
\caption{\label{fig:app} Plots to Table 1 A: Functional data and tMDS
embeddings. Inlier class in grey, outlier class in red. \(r = 0.05\)}
\end{figure}

\begin{figure}
\centering
\includegraphics{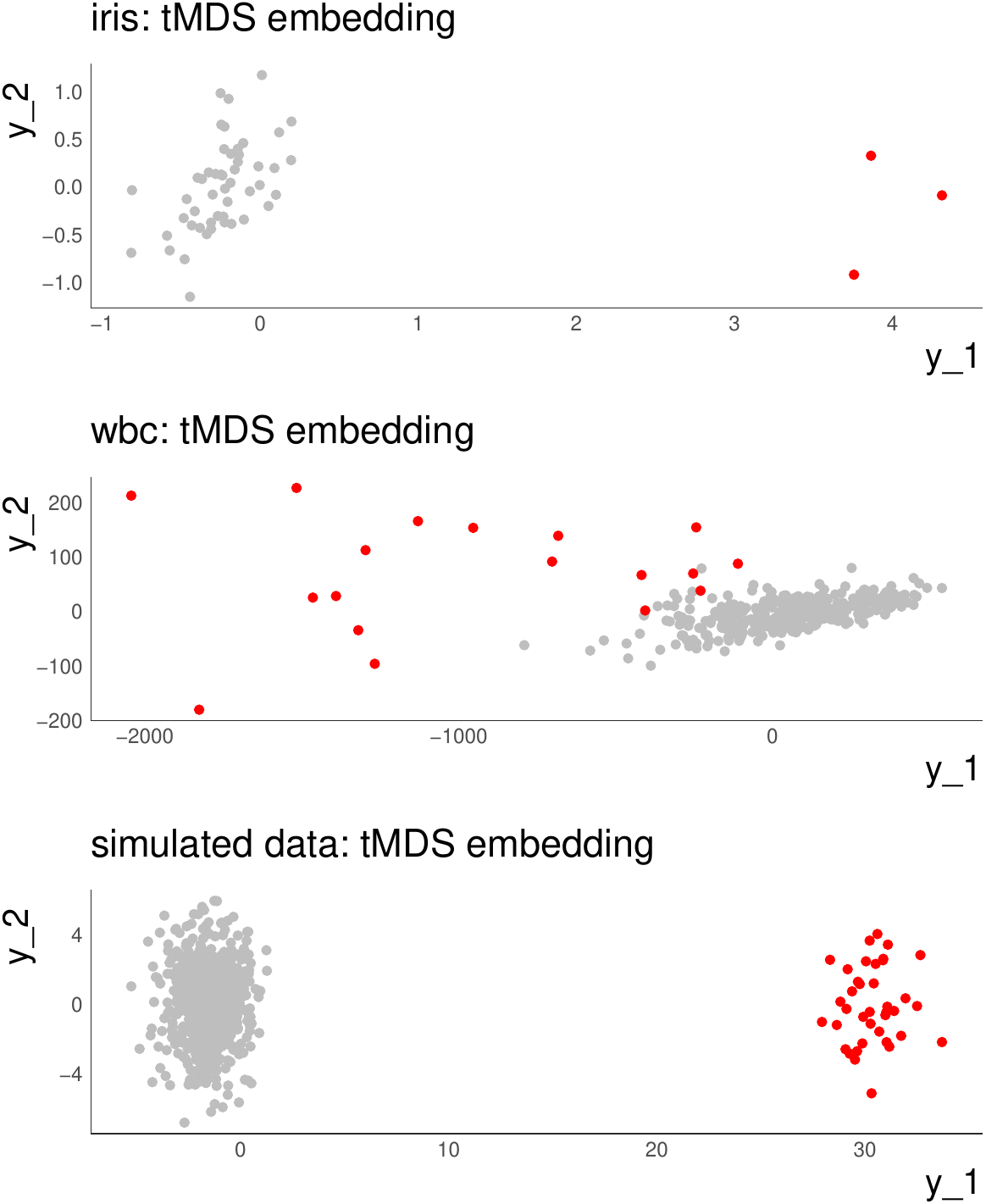}
\caption{\label{fig:app-tab} Plots to Table 1 B: tMDS embeddings of
tabular data. Inlier class in grey, outlier class in red. \(r = 0.05\)}
\end{figure}

\end{document}